\documentclass[11pt,letterpaper]{article}

\usepackage[top=1in, bottom=1in, left=1in, right=1in]{geometry}

\usepackage{microtype}
\usepackage{graphicx}
\usepackage{booktabs} 

\usepackage{amsmath}
\usepackage{amsthm}
\usepackage{amsfonts}
\usepackage{amssymb}
\usepackage{bbm}
\usepackage{mathabx} 
\usepackage{mathrsfs}
\usepackage{color}
\usepackage[inline]{enumitem}
\usepackage{bm}

\usepackage[utf8]{inputenc} 
\usepackage[T1]{fontenc}    
\usepackage{hyperref}       
\usepackage{url}            
\usepackage{booktabs}       
\usepackage{amsfonts}       
\usepackage{nicefrac}       
\usepackage{microtype}      

\usepackage{xcolor}
\usepackage{tikz}
\usepackage[lined]{algorithm2e}
\makeatletter
\renewcommand{\@algocf@capt@plain}{above}
\makeatother
\SetKwInput{kwInput}{Input}
\usepackage{subcaption}

\newcommand{\real}{\mathbb{R}}
\newcommand{\T}{\intercal}
\newcommand{\what}{\hat{w}}
\newcommand{\seq}[1]{[#1]}
\newcommand{\sigmahat}{\hat{\sigma}}
\newcommand{\alphahat}{\hat{\alpha}}
\newcommand{\sign}{\text{sign}}
\newcommand{\prob}{\mathbb{P}}
\newcommand{\E}{\mathbb{E}}
\newcommand{\calO}{\mathcal{O}}

\newtheorem{theorem}{Theorem}
\newtheorem{corollary}{Corollary}
\newtheorem{lemma}{Lemma}

\newtheorem{problem}{Problem Statement}

\title{Exact Support Recovery in Federated Regression with One-shot Communication}
\date{}

\author{%
	Adarsh Barik \\
	Department of Computer Science\\
	Purdue University\\
	West Lafayette, Indiana, USA\\
	\texttt{abarik@purdue.edu} \\
	\and
	Jean Honorio \\
	Department of Computer Science \\
	Purdue University \\
	West Lafayette, Indiana, USA\\
	\texttt{jhonorio@purdue.edu} \\
}

\begin{document}
	
	\maketitle
	
	\begin{abstract}
		Federated learning provides a framework to address the challenges of distributed computing, data ownership and privacy over a large number of distributed clients with low computational and communication capabilities.   In this paper, we study the problem of learning the exact support of sparse linear regression in the federated learning setup. We provide a simple communication efficient algorithm which only needs one-shot communication with the centralized server to compute the exact support. Our method does not require the clients to solve any optimization problem and thus, can be run on devices with low computational capabilities. Our method is naturally robust to the problems of client failure, model poisoning and straggling clients. We formally prove that our method requires a number of samples per client that is polynomial with respect to the support size, but independent of the dimension of the problem. We require the number of distributed clients to be logarithmic in the dimension of the problem. If the predictor variables are mutually independent then the overall sample complexity matches the optimal sample complexity of the non-federated centralized setting. Furthermore, our method is easy to implement and has an overall polynomial time complexity.
	\end{abstract}
	
	\section{Introduction}
	\label{sec:introduction}
	
	Modern day edge devices, with their data acquisition and storage ability, have pushed the need of distributed computing beyond the realms of data centers. Devices such as mobile phones, sensor systems in vehicles, wearable technology and smart homes, within their limited storage and processing capabilities, can constantly collect data and perform simple computations. However, due to data privacy concerns and limitations on network bandwidth and power, it becomes impractical to transmit all the collected data to a centralized server and conduct centralized training. 
	
	The nascent field of federated learning \cite{konevcny2015federated,konevcny2016federated,brendan2017aguera,mohri2019agnostic,li2020federated} tries to address these concerns. As described in \cite{konevcny2016federated}, federated learning is a machine learning setting where the goal is to train a high-quality centralized model with training data distributed over a large number of clients. Unlike the data centers, the clients collect data samples independently but in a non-i.i.d. fashion. The clients may be highly unbalanced, i.e., the number of samples per client may vary significantly. The clients may also have hardware related constraints. Although the number of clients could be quite large,  each client is typically a simple device which has access to a very small number of data samples and can only conduct very basic computations due to limitations on its processing and power capabilities. Furthermore, since battery power is at a premium, the communication between the client and the centralized server acts as a major bottleneck. Due to these constraints, it is common to encounter straggling and faulty clients in the federated learning setting. 
	
	In this work, we study the problem of exact support recovery of sparse linear regression in federated learning. \cite{wainwright2009info} provided an information theoretic lower bound for sparse linear regression. They showed that, in a completely centralized setting where all the data resides in a single server, $\calO(s \log d)$ samples are necessary for exact support recovery of a $d$-dimensional parameter vector with $s$ non-zero entries. In our setting, none of the clients has the access to necessary number of data samples required for exact support recovery or possess computational capabilities to run complex algorithms. Furthermore, we only allow for one-shot communication between the clients and the centralized server, i.e., clients can send information to the centralized server only once. We propose a novel yet simple algorithm for this setting and show that local clients can collaboratively recover the exact support of the sparse linear regression model with provable theoretical guarantees. 
	
	\paragraph{Related work.}
	
	Despite being a new research area, there has been lot of interest in the field of federated learning. On the experimental side, \cite{konevcny2015federated} were the first to formally define federated learning and proposed an algorithm with encouraging experimental results. \cite{konevcny2016federated} came up with strategies to improve the communication efficiency for federated learning. \cite{brendan2017aguera} proposed a communication efficient algorithm for deep networks. Similarly, \cite{yurochkin2019bayesian} developed a novel framework for federated learning with neural networks and \cite{wang2020federated} proposed a federated learning algorithm using matched averaging for neural networks. \cite{bhagoji2019analyzing} empirically analyzed adversarial attacks on federated learning settings. They specifically studied the threat of model poisoning where the adversary controls a small number of malicious clients (usually 1)
	with the aim of causing the global model to misclassify. \cite{li2020fair} studied fair resource allocation in federated learning. On the theoretical side, \cite{he2018cola} proposed a new decentralized training algorithm with guarantees for convergence rates for linear classification and regression models. \cite{smith2017cocoa} presented a communication efficient decentralized framework which covers general non-strongly-convex regularizers, including problems like lasso with convergence rate guarantees. They also describe a possible extension of their method to one-shot communication schemes. \cite{smith2017federated} proposed a multi-task learning based approach for federated learning with convergence rate guarantees which was tolerant to client failure and could handle clients which lag in sending information to the centralized server (also known as straggling clients). \cite{mohri2019agnostic} proposed a client distribution agnostic framework for federated learning. They also provided Rademacher-based generalization bounds for their proposed approach.  
	
	\paragraph{Our Contribution.}
	All the work mentioned above are interesting in their own domain however our contribution is mostly theoretical. The existing theoretical work provide guarantees for convergence rates (which guarantees a small mean squared error in the training set provided enough iterations) or generalization bounds (which guarantees a small mean squared error in the testing set provided enough samples). However, the final solution may not match exactly with the true parameter vector.  In this work, we provide provable theoretical guarantees for exact recovery of the support of the true sparse paramater vector of linear regression in federated learning.
	Support recovery, i.e., correctly detecting the zero and nonzero entries of the parameter vector, is arguably a challenging task. In particular, we show that for a $d$-dimensional $s$-sparse parameter vector $\calO(\log d)$ clients and $\calO(s^2 \log s)$ data samples per client are sufficient to recover the exact support. If the predictor variables are mutually independent then we can do exact support recovery with only $\calO(s)$ data samples per client. Notice that in this case the aggregate sample complexity is $\calO(s\log d)$ which matches the optimal sample complexity of the centralized setting\cite{wainwright2009info,wainwright2009sharp}. We propose a simple yet effective method for exact support recovery and prove that the method is \emph{correct} and efficient in terms of \emph{time} and \emph{sample complexity}. Our method has the following key properties:
	\begin{itemize}
		\item \textbf{Simplicity: } We do not solve any optimization problem at the client level. All the computations are simple and let us use our method in devices with low computational power.
		\item \textbf{One shot communication and privacy: } Our method is communication efficient. We only need one round communication of at most $d-$bits from the client to the centralized server. As the communication is kept to a minimum, very little information about the client is passed to the centralized server. 
		\item \textbf{Fault tolerance and aversion to model poisoning and straggling: } Our method is naturally robust to client node failure and averse to rogue and straggling clients.
	\end{itemize}

	\section{Preliminaries}
	\label{sec:preliminaries}
	In this section, we collect the notation which we use throughout this paper. We also formally define the support recovery problem for sparse linear regression in federated learning.
	
	\subsection{Notation and Problem Setup}
	\label{subsec:notation and problem setup}
	
	Let $w^* \in \real^d$ be a $d-$dimensional parameter with sparsity $s$, i.e., only $s$ out of $d$ entries of $w^*$ are non-zero. We use $\seq{r}$ as a shorthand notation to denote the set $\{1,2,\cdots,r\}$.  Let $S^*$ be the true support set, i.e., $S^* = \{ r | w^*_r \ne 0, r \in \seq{d} \}$. We denote the corresponding complementary non-support set as $S^*_c = \{ r | w^*_r = 0, r \in \seq{d} \}$. Assume there are $g$ clients, each with $n_i$ independent samples, for $i \in [g]$. Note that the data distribution across $g$ clients need not be identical. Each client $i \in \seq{g}$ contains each data sample in the format $(X_i, y_i)$ where $X_i \in \real^d$ are the predictor variables and $y_i \in \real$ is the response variable. The data generation process for each client $i \in \seq{g}$ is as follows:
	\begin{align}
	\label{eq:generative model}
	\begin{split}
	y_i = X_i^\T w^* + e_i \; ,
	\end{split}
	\end{align} 
	where $e_i$ is a zero mean sub-Gaussian additive noise with variance proxy $\eta_i^2$, where $\eta_i > 0$. Note that all the clients share the same parameter vector $w^*$. The $j$-th entry of $X_i$ is denoted by $X_{ij}, \forall i\in \seq{g}, j \in \seq{d}$.  Each entry $X_{ij}$ of $X_i$ is a zero mean sub-Gaussian random variable with variance proxy $\rho_i^2$, where $\rho_i > 0$. We denote covariance matrix for $X_i$ as $\Sigma^i \in \real^{d \times d}$ with diagonal entries $\Sigma^i_{jj} \equiv {\sigma^i_{jj}}^2, \forall j \in \seq{d}$ and non-diagonal entries $\Sigma^i_{jk} \equiv \sigma^i_{jk}, \forall j,k \in \seq{d}, j \ne k$. If predictor variables are mutually independent then $\sigma^i_{jk} = 0, \forall i \in \seq{g}, j,k \in \seq{d}, j \ne k$. The $t$-th sample of the $i$-th client is denoted by $(X_i^t, y_i^t), \forall i \in \seq{g}, t \in \seq{n_i}$. We note that $X_i^t \in \real^d$ and $y_i^t \in \real$ and denote $j$-th entry of $X_i^t$ as $X_{ij}^t$. Notice that the data distributions for $(X_i, y_i)$ can vary a lot across the clients by varying $\rho_i$ and $\eta_i$, as well as the specific sub-Gaussian probability distribution. The class of sub- Gaussian variates includes for instance Gaussian variables, any bounded random variable (e.g., Bernoulli, multinomial, uniform), any random variable with strictly log-concave density, and any finite mixture of sub-Gaussian variables. Similarly, data samples can be distributed unevenly across the clients by varying $n_i$. In subsequent sections, we use $\prob(A)$ to denote probability of the event $A$ and $\E(A)$ to denote the expectation of the random variable $A$.
	
	\subsection{Problem Statement}
	\label{subsec: problem statement}
	
	For our problem, we assume that $n_i < \calO(s \log d), \forall i \in \seq{g}$. Otherwise, support can be trivially recovered by using compressed sensing methods in the client with $n_i = \calO(s \log d)$ which is the order of necessary and sufficient number of samples required for exact support recovery in linear regression setup \cite{wainwright2009info,wainwright2009sharp}. Furthermore, we assume that each of our clients can only do very simple computations and can only do one-shot communication with the centralized server, i.e., each client can only send at most $d$-bits to the centralized server. Considering the above requirements, we are interested in answering the following question:
	\begin{problem}[Exact Support Recovery]
		\label{prob:exact support recovery}
		Given that each client contains $n_i < \calO(s\log d)$ data samples generated through the process described in equation \eqref{eq:generative model}, is it possible to efficiently recover the true support of the $s$-sparse shared parameter vector $w^* \in \real^d$ by collecting $d$-bits of information from every client only once with provable theoretical guarantees.
	\end{problem}
	The efficiency in exact recovery means that the sample complexity per client should be strictly less than $\calO(s\log d)$ and that our algorithm should have polynomial time complexity and should also be easy to implement.    
	
	\section{Our Method}
	\label{sec:methodology}
	
	In this section, we present a simple algorithm to solve problem \ref{prob:exact support recovery}. Our main idea is that estimation at the client level can be incorrect for every client but this information can still be aggregated in a careful manner to compute the true support.    
	
	\subsection{Client Level Computations}
	\label{subsec:client level computations}
	
	Each client tries to estimate the support of $w^*$ using $n_i$ independent samples available to it. As mentioned previously, $n_i, \forall i \in \seq{g}$ is not sufficient to compute correct support of $w^*$ using any method possible \cite{wainwright2009info}. Let $\what_i \in \real^d$ be the estimate of $w^*$ computed by each client $i$. Let $S_i$ be the support of $\what_i$. Each server communicates the computed support (at most $d$ bits) to a centralized server which then computes the final support for $w^*$. The centralized server receives $S_i$ from each client and computes the final support $S = f(S_1, S_2,\cdots,S_g)$. Each client $i, \forall i \in \seq{g}$ computes $\what_i$ in the following way:
	\begin{align}
	\label{eq:what}
	\begin{split}
	\forall i \in \seq{g}, j \in \seq{d},\quad \what_{ij} = \frac{1}{\sigmahat_{ij}} \sign(\alphahat_{ij}) \max (0, |\alphahat_{ij}| - \lambda) \; ,
	\end{split}
	\end{align} 
	where $\what_{ij}$ is $j$-th entry of $\what_i$ and $\lambda > 0$ is a regularization parameter. We present the exact procedure to compute a feasible $\lambda$ in later sections. We also define $\sigmahat_{ij}$ and $\alphahat_{ij}$ as follows: 
	\begin{align}
	\label{eq:sigma_alpha}
	\begin{split}
	\sigmahat_{ij} \triangleq \frac{1}{n_i} \sum_{t=1}^{n_i} (X_{ij}^t)^2,\quad \alphahat_{ij} \triangleq \frac{1}{n_i} \sum_{t=1}^{n_i} y_i^t X_{ij}^t
	\end{split}
	\end{align}
	Note that these are simple calculations and can be done in $\calO(d n_i)$ run time at each client. If $n_i$ can be kept small (which we will show later), this can be done even by a device with low computational ability. The choice of this exact form of $\what_{ij}$ in equation \eqref{eq:what} is not arbitrary. To get the intuition behind our choice, consider the following $\ell_1$-regularized (sparse) linear regression problem at each client.
	\begin{align}
	\label{eq:serverlasso}
	\begin{split}
	(\forall i\in\seq{g}), \quad \what_i = \arg\min_w \frac{1}{n_i}\sum_{t=1}^{n_i} (w^\T X_i^t - y_i^t)^2 + \lambda \| w \|_1 \; ,
	\end{split}
	\end{align}
	where $\| \cdot \|_1$ denotes the $\ell_1$ norm of a vector. The construction of $\what_i$ in equation \eqref{eq:what} is the exact solution to optimization problem \eqref{eq:serverlasso} if predictor variables, i.e., the entries in $X_{i}$, are assumed to be uncorrelated. Notice how the solution provided in \eqref{eq:what} avoids any computation (or estimation) of the covariance matrix which, in any case, would be incorrect if each client has access to only a few samples. Each client $i$ sends the support $S_i = \{ j | \what_{ij} \ne 0, j \in \seq{d} \}$ of $\what_i$ to the centralized server. Note that even in the worst case scenario, each client only sends $d$ bits to the centralized server.
	
	\subsection{Information Aggregation and Constructing the Final Support}
	\label{subsec:constructing final support S}
	
	We aggregate supports $S_i, \forall i \in \seq{g}$ from all the clients and construct the final support. Before we get to the construction of the final support, we define a random variable $R_{ij}, \forall i \in \seq{g}, j \in \seq{d}$ which takes value $1$ if $j \in S_i$ and $0$ otherwise.

	Thus, random variable $R_{ij}$ indicates whether entry $j$ is in the support $S_i$ of client $i$. Using the random variables $R_{ij}$, we construct the final support $S$ by computing the median of $R_{ij}$ across $i \in \seq{g}$. If the median is $1$ then we conclude that $j$ is in the support otherwise we conclude that $j$ is not in the support. More formally, we define a random variable $R_j \triangleq \frac{1}{g}\sum_{i=1}^g R_{ij}$ and if $R_j \geq \frac{1}{2}$, then we conclude that $j \in S$. Otherwise, if $R_j < \frac{1}{2}$, then we conclude that $j \notin S$. The above procedure can be compactly written as the following algorithms running in clients and centralized server:
	
	\begin{algorithm}[H]
		\label{algo:getExactSupport}
		\begin{minipage}{0.5\textwidth}
			\begin{algorithm}[H]
				\SetKwInOut{Input}{Input}
				\SetKwInOut{Output}{Output}
				\tcp{Runs in client $i, \forall i \in \seq{g}$} 
				\Input{Data samples $(X_i^t, y_i^t), \forall t \in \seq{n_i}$ }
				\Output{Estimated support for shared parameter $w^*$}
				$R_i \leftarrow \{0\}^d$ \;
				\For{each $j \in \seq{d}$}{
					Compute $\what_{ij}$ using equation \eqref{eq:what} and \eqref{eq:sigma_alpha} \;
					\If{$\what_{ij} \ne 0$}{ 
						$R_{ij} \leftarrow 1 $ \;
					}
				}
				Send $R_i$ to centralized server \;
			\end{algorithm}
		\end{minipage}%
		\begin{minipage}{0.5\textwidth}
			\vspace{-1.8\baselineskip}
			\begin{algorithm}[H]
				\SetKwInOut{Input}{Input}
				\SetKwInOut{Output}{Output}
				\tcp{Runs in centralized server}
				\Input{$R_i, \forall i \in \seq{g}$}
				\Output{True support $S$ for shared parameter $w^*$}
				$S \leftarrow \{\}$ \;
				\For{each $j \in \seq{d}$}{ 
					Compute $R_j = \frac{1}{g} \sum_{i=1}^g R_{ij}$ \;
					\If{$R_j \geq \frac{1}{2}$}{
						$S \leftarrow S \cup \{j\}$ \;
					}
				}
			\end{algorithm}
		\end{minipage}
		\caption{getExactSupport}
	\end{algorithm}

	\section{Main Results and Analysis}
	\label{sec:analysis}
	
	In this section, we describe and analyze our theoretical results. We present our results in two different settings. In the first setting, we assume that predictor variables are mutually independent. We tackle the more general case of correlated predictors in the second setting.
	
	\subsection{Mutually Independent Predictors}
	\label{subsec:mutually independent predictors}
	
	In this setting, predictor variables are mutually independent of each other in all the clients, i.e., $\forall i \in \seq{g}$, $\E(X_{ij} X_{ik}) = 0, \forall j, k \in \seq{d}, j \ne k$. In this setting, we state the following result:
	\begin{theorem}[Mutually Independent Predictors]
		\label{thm:mutually independent predictors}
		For the federated support learning for linear regression as described in Section \ref{sec:preliminaries} where predictor variables are mutually independent of each other, if for some $0 < \delta < 1$, each of the $g = \calO(\log d)$ client has $n_i = \calO(\frac{1}{\delta^2})$ data samples and if for each $i \in \seq{g}$ and $j \in S^*$,
		\begin{align*}
		\begin{split}
		8 \delta \rho_i^2 \sqrt{\sum_{k\in S^*} w_k^2} + 8 |\eta_i\rho_i|\delta < \lambda < |w_j^* {\sigma_{jj}^i}^2| - 8 |w_j^*| \rho_i^2 \delta - 8 \rho_i^2 \sqrt{\sum_{k\in S^*, k \ne j} {w_k^*}^2} \delta - 8 |\eta_i\rho_i| \delta \;,
		\end{split}
		\end{align*} 
		then Algorithm \ref{algo:getExactSupport} recovers the exact support for the shared parameter vector $w^*$.   
	\end{theorem}  
	By taking $\delta = \calO(\frac{1}{\sqrt{s}})$, we get the following corollary:
	\begin{corollary}
		\label{cor:mutually independent predictors}
		For the federated support learning for linear regression as described in Section \ref{sec:preliminaries} where predictor variables are mutually independent of each other, if for some constant $0 < K < \sqrt{s}$, each of the $g = \calO(\log d)$ client has $n_i = \calO(s)$ data samples and if for each $i \in \seq{g}$ and $j \in S^*$,
		\begin{align*}
		\begin{split}
		&8 K \rho_i^2 \sqrt{\frac{\sum_{k\in S^*} {w_k^*}^2}{s}} + 8 K \frac{|\eta_i\rho_i|}{\sqrt{s}} < \lambda < |w_j^* {\sigma_{jj}^i}^2| - 8 K \frac{|w_j^*| \rho_i^2}{\sqrt{s}}  - 8 K \rho_i^2 \sqrt{\frac{\sum_{k\in S^*, k \ne j} {w_k^*}^2}{s}}  - \\
		& 8 K \frac{|\eta_i\rho_i|}{\sqrt{s}} \;,
		\end{split}
		\end{align*} 
		then Algorithm \ref{algo:getExactSupport} recovers the exact support for the shared parameter vector $w^*$.
	\end{corollary}
	The choice of such value of $\delta$ is to subdue the growth of the $\sqrt{\sum_{k\in S^*} {w_k^*}^2}$ term which approximately grows as $\calO(\sqrt{s})$. Later on, we will empirically show that such a choice leads to a feasible range for $\lambda$. Also observe that, the overall sample complexity for our algorithm is $\calO(s \log d)$ which matches the optimal sample complexity for sparse linear regression\cite{wainwright2009info,wainwright2009sharp}, i.e., even if we have access to all the samples in a centralized server, we can not have a better sample complexity guarantee for support recovery. 
	
	\subsubsection{Proof of Theorem \ref{thm:mutually independent predictors}}
	\label{subsubsec: proof of theorem mutually independent predictors}
	
	\begin{proof}
		\label{proof:theorem mutually independent predictors}
		Recall that $R_j = \frac{1}{g} \sum_{i=1}^g R_{ij}$ where $R_{ij}$ is defined in Section \ref{subsec:constructing final support S}. We prove that, with high probability,  $R_j \geq \frac{1}{2}, \forall j \in S^*$ and $R_j < \frac{1}{2}, \forall j \in S^*_c$. We will provide the proof in two parts. First, we deal with entries $j$ which are in the support of $w^*$, i.e., $j \in S^*$ and then we will deal with $j \in S^*_c$.
		
		\paragraph{For entries $j$ in support $S^*$.}
		
		We begin our proof by first stating the following lemma.
		\begin{lemma}
			\label{lem:support mcdiarmid}
			For $j \in S^*$, let $\E(R_j) > \frac{1}{2}$, then $R_j \geq \frac{1}{2}$ with probability at least $1 - 2 \exp( - 2g (-\frac{1}{2} + \E(R_j))^2)$.
		\end{lemma}
		
		Next we show that for $j \in S^*$, $\E(R_j)$ is indeed greater than $\frac{1}{2}$. To that end, we provide the result of the following lemma.
		\begin{lemma}
			\label{lem:support bound on E(Rj)}
			For $j \in S^*$ and some $0 < \delta \leq 1$, if predictors are mutually independent of each other and $ 0 < \lambda < |w_j^* {\sigma_{jj}^i}^2| - 8 |w_j^*| \rho_i^2 \delta - 8 \rho_i^2 \sqrt{\sum_{k\in S^*, k \ne j} {w_k^*}^2} \delta - 8 |\eta_i\rho_i| \delta $
			then $ \E(R_j) \geq 1 - \frac{6}{g} \sum_{i=1}^g \exp(-n_i\delta^2) $. Furthermore, for $n_i = \calO(\frac{1}{\delta^2})$, we have $\E(R_j) > \frac{1}{2}$.
		\end{lemma}
		
		\paragraph{For entries $j$ in non-support $S^*_c$.}
		Similar to the entries in the support, we begin this part by stating the following result for entries in the non-support.
		\begin{lemma}
			\label{lem:nonsupport mcdiarmid}
			For $j \in S^*_c$, let $\E(R_j) < \frac{1}{2}$, then $R_j \leq \frac{1}{2}$ with probability at least $1 - 2 \exp( - 2g (\frac{1}{2} - \E(R_j))^2)$.
		\end{lemma}
		It remains to show that for $j \in S^*_c$, $\E(R_j)$ is smaller than $\frac{1}{2}$. In particular, we use the result from the following lemma.
		\begin{lemma}
			\label{lem:nonsupport bound E(Rj)}
			For $j \in S^*_c$ and $0 < \delta \leq 1$, if predictors are mutually independent of each other and if $ \lambda > 8 \delta \rho_i^2 \sqrt{\sum_{k\in S^*} w_k^2} + 8 |\eta_i\rho_i|\delta $
			then $\E(R_j) \leq \frac{4}{g} \sum_{i=1}^g \exp(-n_i \delta^2)$. Furthermore, for $n_i = \calO(\frac{1}{\delta^2})$, we have $\E(R_j) < \frac{1}{2}$.
		\end{lemma} 
		Results from Lemma \ref{lem:support bound on E(Rj)} and \ref{lem:nonsupport bound E(Rj)} make sure that Lemma \ref{lem:support mcdiarmid} and \ref{lem:nonsupport mcdiarmid} hold. We would like these results to hold across all $j \in \seq{d}$. This implies that we need a union bound across all the $d$ predictors. Thus, having $g = \calO(\log d)$ ensures that our results hold for all entries in the support and non-support with high probability.  
	\end{proof}
	
	\subsection{Correlated predictors}
	\label{subsec:correlated predictors}
	
	Now that we have dealt with mutually independent predictors, we focus on correlated predictors in this section. As described previously, the covariance matrix for $X_i$ is denoted by $\Sigma^i \in \real^{d \times d}$ with diagonal entries $\Sigma^i_{jj} \equiv {\sigma^i_{jj}}^2, \forall j \in \seq{d}$ and non-diagonal entries $\Sigma^i_{jk} \equiv \sigma^i_{jk}, \forall j,k \in \seq{d}, j \ne k$. Some of the results from the previous subsection can be used  for this setting as well. However, correlation between predictors affects some results. Below, we state the main results for this setting before proving them formally.  
	\begin{theorem}[Correlated Predictors]
		\label{thm:correlated predictors}
		For the federated support learning for linear regression as described in Section \ref{sec:preliminaries}, if for some $0 < \delta < \frac{1}{\sqrt{2}}$, each of the $g = \calO(\log d)$ client has $n_i = \calO(\frac{1}{\delta^2} \log s)$ data samples and if for each $i \in \seq{g}$,
		\begin{align*}
		\begin{split}
		&(\forall j \in S^*_c) |\sum_{k\in S^*} w^*_k \sigma^i_{jk}| + \sum_{k\in S^*} 8 \sqrt{2} |w_k^*| (1 + 4 \max_j \frac{\rho_i^2}{{\sigma^i_{jj}}^2}) \max_j {\sigma^i_{jj}}^2 \delta + 8 |\eta_i \rho_i| \delta < \lambda \\
		&< (\forall j \in S^*) |(w^*_j{\sigma^i_{jj}}^2 + \sum_{k\in S^*, k\ne j} w^*_k \sigma^i_{jk})| - 8 |w_j^*| \rho_i^2 \delta - \sum_{k\in S^*, k\ne j} 8 \sqrt{2} |w_k^*| (1 + 4 \max_j \frac{\rho_i^2}{{\sigma^i_{jj}}^2}) \\
		&\max_j {\sigma^i_{jj}}^2 \delta - 8 |\eta_i \rho_i| \delta \;,
		\end{split}
		\end{align*} 
		then Algorithm \ref{algo:getExactSupport} recovers the exact support for the shared parameter vector $w^*$.   
	\end{theorem}  
	By taking $\delta = \calO(\frac{1}{s})$, we get the following corollary:
	\begin{corollary}
		\label{cor:correlated predictors}
		For the federated support learning for linear regression as described in Section \ref{sec:preliminaries}, if for some constant $0 < K < \frac{s}{\sqrt{2}}$, each of the $g = \calO(\log d)$ client has $n_i = \calO(s^2 \log s)$ data samples and if for each $i \in \seq{g}$,
		\begin{align*}
		\begin{split}
		&(\forall j \in S^*_c) |\sum_{k\in S^*} w^*_k \sigma^i_{jk}| + \sum_{k\in S^*} 8 \sqrt{2} |w_k^*| (1 + 4 \max_j \frac{\rho_i^2}{{\sigma^i_{jj}}^2}) \max_j {\sigma^i_{jj}}^2 \frac{K}{s} + 8 |\eta_i \rho_i| \frac{K}{s} < \lambda \\
		&< (\forall j \in S^*) |(w^*_j{\sigma^i_{jj}}^2 + \sum_{k\in S^*, k\ne j} w^*_k \sigma^i_{jk})| - 8 |w_j^*| \rho_i^2 \frac{K}{s} - \sum_{k\in S^*, k\ne j} 8 \sqrt{2} |w_k^*| (1 + 4 \max_j \frac{\rho_i^2}{{\sigma^i_{jj}}^2}) \\
		&\max_j {\sigma^i_{jj}}^2 \delta - 8 |\eta_i \rho_i| \frac{K}{s} \;,
		\end{split}
		\end{align*} 
		then Algorithm \ref{algo:getExactSupport} recovers the exact support for the shared parameter vector $w^*$.
	\end{corollary}
	
	As with the previous case, the choice of such a value of $\delta$ is to subdue the growth of terms which grow as $\calO(s)$. In our experiments, this leads to a feasible range for $\lambda$. In this case, the overall sample complexity for our algorithm is $\calO(s^2 \log s \log d)$ which only differs by a factor of $s\log s$ from the optimal sample complexity for support recovery in sparse linear regression in the centralized setting where all the data resides in a single server\cite{wainwright2009info,wainwright2009sharp}.
	
	\subsubsection{Proof of Theorem \ref{thm:correlated predictors}}
	\label{subsubsec:proof of theorem correlared predictors}
	
	\begin{proof}
		\label{proof:theorem correlared predictors}
		Recall that $R_j = \frac{1}{g} \sum_{i=1}^g R_{ij}$ where $R_{ij}$ is defined in Section \ref{subsec:constructing final support S}. We will again prove that, with high probability,  $R_j \geq \frac{1}{2}, \forall j \in S^*$ and $R_j < \frac{1}{2}, \forall j \in S^*_c$. Some of the results from the previous Section \ref{proof:theorem mutually independent predictors} follow without any changes. We provide new results for the remaining parts. Like before first, we deal with entries $j$ which are in the support of $w^*$, i.e., $j \in S^*$ and then we will deal with $j \in S^*_c$. 
		
		\paragraph{For entries $j$ in support $S^*$.}
		
		We observe that Lemma \ref{lem:support mcdiarmid} holds even in this case. Thus, we start our proof by stating the following lemma.
		\begin{lemma}
			\label{lem:support bound on E(Rj) correlated}
			For $j \in S^*$ and some $0 < \delta \leq \frac{1}{\sqrt{2}}$, if $\forall j \in S^*$,
			$ 0 < \lambda < |(w^*_j{\sigma^i_{jj}}^2 + \sum_{k\in S^*, k\ne j} w^*_k \sigma^i_{jk})| - 8 |w_j^*| \rho_i^2 \delta - \sum_{k\in S^*, k\ne j} 8 \sqrt{2} |w_k^*| (1 + 4 \max_j \frac{\rho_i^2}{{\sigma^i_{jj}}^2}) \max_j {\sigma^i_{jj}}^2 \delta - 8 |\eta_i \rho_i| \delta $
			then $ \E(R_j) \geq 1 - \frac{4s}{g} \sum_{i=1}^g \exp(-n_i\delta^2 ) $. Furthermore, for $n_i = \calO(\frac{1}{\delta^2}\log s)$, we have $\E(R_j) > \frac{1}{2}$.
		\end{lemma}
		
		\paragraph{For entries $j$ in non-support $S^*_c$.}
		Again, Lemma \ref{lem:nonsupport bound E(Rj)} follows directly. Thus, we present the following lemma to show that for the entries in the non-support $\E(R_j) < \frac{1}{2}$.
		\begin{lemma}
			\label{lem:nonsupport bound E(Rj) correlated}
			For $j \in S^*_c$ and some $0 < \delta \leq \frac{1}{\sqrt{2}}$, if
			$ \lambda > |\sum_{k\in S^*} w^*_k \sigma^i_{jk}| + \sum_{k\in S^*} 8 \sqrt{2} |w_k^*| (1 + 4 \max_j \frac{\rho_i^2}{{\sigma^i_{jj}}^2}) \max_j {\sigma^i_{jj}}^2 \delta + 8 |\eta_i \rho_i| \delta $ 
			then $\E(R_j) \leq \frac{4s + 2}{g} \sum_{i=1}^g \exp(-n_i \delta^2)$. Furthermore, for $n_i = \calO(\frac{1}{\delta^2} \log s)$, we have $\E(R_j) < \frac{1}{2}$.
		\end{lemma} 
		Results from Lemmas \ref{lem:support bound on E(Rj) correlated} and \ref{lem:nonsupport bound E(Rj) correlated} ensure that Lemma \ref{lem:support mcdiarmid} and \ref{lem:nonsupport mcdiarmid} hold. Since we would like these results to hold across all $j \in \seq{d}$, we need a union bound across all the $d$ predictors. Thus, having $g = \calO(\log d)$ makes sure that our results hold for all entries in the support and non-support with high probability.  	
	\end{proof}
	
	\subsection{Time Complexity}
	\label{sub:time complexity}
	
	Each client does $\calO(dn_i)$ basic calculations. Thus, from the results of Corollaries \ref{cor:mutually independent predictors} and \ref{cor:correlated predictors}, the time complexity at each client is $\calO(sd)$ for mutually independent predictors and $\calO(s^2 d \log s)$ for correlated predictors. The centralized server gathers $d$-bits of information from $g$ clients in $\calO(dg)$ time.  
	
	\section{Discussion on Robustness}
	\label{sec:discussion on robustness}
	
	Since our method only relies on the correct calculation of the median, it is naturally robust to failure of few clients. To simulate the effect of model poisoning \cite{bhagoji2019analyzing} and stragglers, we consider that $0 < \beta < \frac{1}{2}$ portion of clients have gone rogue (are straggling) and transmitting the wrong information to the centralized server. For the worst case scenario, we assume that they report the complement of the support, i.e., they always send a bit ``$1$'' for entries in the non-support and a bit ``$0$'' for entries in the support. To accommodate this change in the case of correlated predictors, we slightly change statements of Lemmas \ref{lem:support bound on E(Rj) correlated} and \ref{lem:nonsupport bound E(Rj) correlated}. Now we have, $ (\forall j \in S^*), \quad \E(R_j) \geq (1 - \beta) - \frac{4s}{g} \sum_{i=1}^{(1-\beta)g} \exp(-n_i \delta^2) $ and $ (\forall j \in S^*_c), \quad \E(R_j) \leq \frac{4s + 2}{g} \sum_{i=1}^{(1 - \beta)g} \exp(-n_i \delta^2) + \beta $.
	It is easy to see that, as long as, we have $n_i > \frac{1}{\delta^2} \log(\frac{(8s + 4)(1 - \beta)}{1 - 2 \beta })$ data samples per client, then we still have $\E(R_j) > \frac{1}{2}, \forall j \in S^*$ and $\E(R_j) < \frac{1}{2}, \forall j \in S^*_c$ and all our results still hold. A similar analysis can be conducted for the case of mutually independent predictors and our results hold as long as we have $n_i > \frac{1}{\delta^2} \log (\frac{12(1 - \beta)}{1 - 2 \beta})$ data samples per client.       
	
	\section{Experimental Results}
	\label{sec:experimental results}
	
	\begin{figure}[!ht]
		\centering
		\begin{subfigure}{.45\textwidth}
			\centering
			\includegraphics[width=\linewidth]{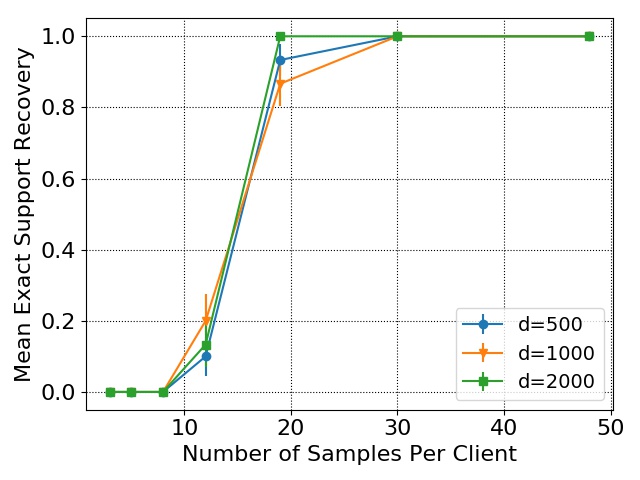}
			\caption{Exact support recovery against numbers of samples per client}
			\label{fig:recnumsample}
		\end{subfigure}%
		\begin{subfigure}{.45\textwidth}
			\centering
			\includegraphics[width=\linewidth]{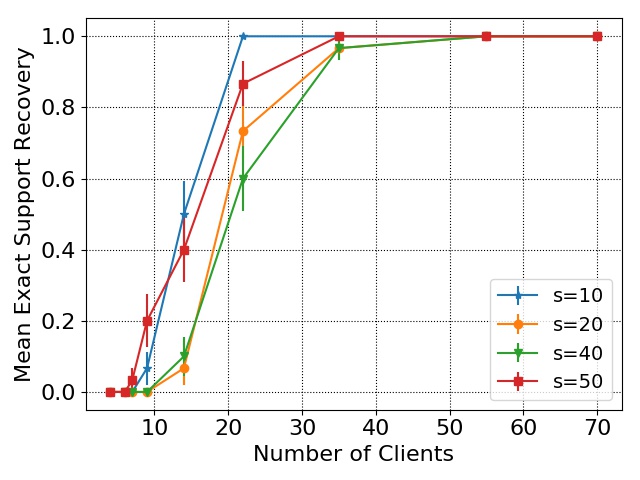}	
			\caption{Exact support recovery against numbers of clients}
			\label{fig:recnumclient}
		\end{subfigure}
		\caption{Phase transition curves. Left: Exact support recovery averaged across $30$ runs against varying number of samples per client for $d = 500, 1000$, and $2000$, $s = 3$, $g = \calO(\log d)$ clients. Right: Exact support recovery averaged across $30$ runs against varying number of clients  for $s = 10, 20, 40$, and $50$, $d = 1000$, $n = \max(30, \calO(s^2 \log s))$ samples per server.}
		\label{fig:recovery}
	\end{figure}

	In this section, we validate our theoretical results by conducting computational experiments. We provide the results for the experiments when predictors are correlated. Data in each client is generated by following generative process described in equation \ref{eq:generative model}. Note that predictors and error term in different clients follow different sub-Gaussian distributions. To make it more general, we keep the correlation between entries in the support different than the correlation between one entry in the support and the other entry in the non-support and these further vary across clients. The regularization parameter $\lambda$ is chosen such that condition in corollary \ref{cor:correlated predictors} is satisfied for every client and for every entry in support and non-support. All the results reported here are averaged over 30 independent runs. We conduct two separate experiments to verify that $n_i = \calO(s^2\log s)$ independent samples per client and a total of $g = \calO(\log d)$ clients are sufficient to recover the true support.
	
	\paragraph{Exact support recovery against number of samples per client.}  
	This experiment was conducted for a varying number of predictors ($d = 500, 1000$ and $2000$). For each of them, we fixed the number of clients to be $g = 2 \log d$. The sparsity $s$ is kept fixed at $3$. The number of samples per client $n_i$ is varied with control parameter $C$ as $10^C s^2 \log s$. Performance of our method is measured by assigning value $1$ for exact recovery and $0$ otherwise. We can see in Figure \ref{fig:recnumsample}, that initially, recovery remains at $0$ and then there is sharp jump after which recovery becomes $1$. Notice how all three curves align perfectly. This validates the result of our theorem and shows that given $g = \calO(\log d)$ clients, $n_i = \calO(s^2 \log s)$ samples per client are sufficient to recover the true support. 
	
	\paragraph{Exact support recovery against number of clients.}
	The second experiment was conducted for a varying number of non-zero entries ($s=10, 20,  40$ and $50$) in the support of $w^*$. The experiments were run for a setup with $d=1000$ predictors. We fixed the number of samples per client ($n_i$) to be $\max(30, \calO(s^2 \log s))$. This ensures that a minimum of $30$ samples are available to each client. This is inline with our previous experiment where exact recovery is achieved around $30$ samples per client. The number of clients $g$ is varied with control parameter $C$ as $10^C \log d$. Like previous experiment, performance is measured by assigning value $1$ for exact recovery and $0$ otherwise. We can again see in Figure \ref{fig:recnumclient}, that initially, recovery is remains at $0$ and then it goes to $1$ as we increase number of clients. We also notice that all four curves align nicely. This validates that given $n_i = \calO(s^2 \log s)$ independent samples per server,  $g = \calO(\log d)$ clients are sufficient to recover the true support.  

	\section{Concluding Remarks}
	\label{sec:conclusion}
	In this paper, we propose a simple and easy to implement method for learning the exact support of parameter vector of linear regression problem in a federated learning setup. We provide theoretical guarantees for the correctness of our method. We also show that our method runs in polynomial sample and time complexity. Furthermore, our method is averse to client failures, model poisoning and straggling clients. As a future direction, it would be interesting to analyze if the bound on the sample complexity in the case of correlated predictors matches corresponding information theoretic lower bounds.


	\pagebreak
	\appendix

	\section{Proof of Lemma \ref{lem:support mcdiarmid}}
	\label{sec:proof of lemma support mcdiarmid} 
	\emph{\paragraph{Lemma \ref{lem:support mcdiarmid}}
		For $j \in S^*$, let $\E(R_j) > \frac{1}{2}$, then $R_j \geq \frac{1}{2}$ with probability at least $1 - 2 \exp( - 2g (-\frac{1}{2} + \E(R_j))^2)$.
	}
	\begin{proof}
		\label{proof:support mcdiarmid}
		Note that since $w^*$ is fixed, for any $j \in \seq{d}$, $R_{ij}$ across  $i \in \seq{g}$ are independent of each other. 
		\begin{align}
		\begin{split}
		\prob(R_j < \frac{1}{2}) &= \prob(R_j - \E(R_j) < \frac{1}{2} - \E(R_j)) \\
		&\leq \prob(|R_j - \E(R_j)| > - \frac{1}{2} + \E(R_j))
		\end{split}
		\end{align}
		The last inequality holds as long as $\E(R_j) > \frac{1}{2}$. We also note that for a fixed $w^*$, $R_j = f(R_{1j}, R_{2j}, \cdots, R_{gj})$ such that 
		\begin{align}
		\begin{split}
		(\forall k \in \seq{g}),\quad \sup_{R_{1j}, \cdots, R_{kj}, R_{kj}',\cdots, R_{gj}}|f(R_{1j}, \cdots, R_{kj},\cdots, R_{gj}) - f(R_{1j}, \cdots, R_{kj}',\cdots, R_{gj})| \leq \frac{1}{g}
		\end{split}
		\end{align}
		Thus using McDiarmid's inequality \cite{mcdiarmid1989method}, we can write the following:
		\begin{align}
		\begin{split}
		\prob(|R_j - \E(R_j)| > - \frac{1}{2} + \E(R_j)) \leq 2 \exp( - 2g (-\frac{1}{2} + \E(R_j))^2)
		\end{split}
		\end{align}
	\end{proof}
	
	\section{Proof of Lemma \ref{lem:support bound on E(Rj)}}
	\label{sec:proof of lemma lem:support bound on E(Rj)}
	\emph{\paragraph{Lemma \ref{lem:support bound on E(Rj)}}
		For any $j \in S^*$ and some $0 < \delta \leq 1$, if predictors are mutually independent of each other and
		\begin{align}
		\begin{split}
		0 < \lambda < |w_j^* {\sigma_{jj}^i}^2| - 8 |w_j^*| \rho_i^2 \delta - 8 \rho_i^2 \sqrt{\sum_{k\in S^*, k \ne j} {w_k^*}^2} \delta - 8 |\eta_i\rho_i| \delta
		\end{split}
		\end{align} 
		then $ \E(R_j) \geq 1 - \frac{6}{g} \sum_{i=1}^g \exp(-n_i\delta^2) $. Furthermore, for $n_i = \calO(\frac{1}{\delta^2})$, we have $\E(R_j) > \frac{1}{2}$.
	}
	\begin{proof}
		\label{proof:support bound on E(Rj)}
		By using sum of expectations, we can write $\E(R_j)$ in the following form: 
		\begin{align}
		\begin{split}
		\E(R_j) &= \frac{1}{g} \sum_{i=1}^g \E(R_{ij}) \\
		&= \frac{1}{g} \sum_{i=1}^g \prob(\what_{ij} \ne 0) \\
		&= \frac{1}{g} \sum_{i=1}^g \prob( \frac{1}{\sigmahat_{ij}} \sign(\alphahat_{ij}) \max (0, |\alphahat_{ij}| - \lambda)  \ne 0) \\
		&= \frac{1}{g} \sum_{i=1}^g \prob(\sign(\alphahat_{ij}) \ne 0 \wedge |\alphahat_{ij}| > \lambda) \\
		&= \frac{1}{g} \sum_{i=1}^g \prob( \alphahat_{ij} \ne 0 \mid |\alphahat_{ij}| > \lambda ) \prob( |\alphahat_{ij}| > \lambda ) \\
		&= \frac{1}{g} \sum_{i=1}^g \prob( |\alphahat_{ij}| > \lambda )
		\end{split}
		\end{align}
		The last equality follows since for $\lambda > 0$, we have $\prob( \alphahat_{ij} \ne 0 \mid |\alphahat_{ij}| > \lambda ) = 1$. Now, we put a bound on the term $\prob( |\alphahat_{ij}| > \lambda )$.
		\begin{align}
		\begin{split}
		\prob( |\alphahat_{ij}| > \lambda ) &= \prob( |\frac{1}{n_i} \sum_{t=1}^{n_i} y_i^t X_{ij}^t| > \lambda ) \\
		& \text{Expanding $y_i^t$ using equation \eqref{eq:generative model}} \\
		&= \prob( |\frac{1}{n_i} \sum_{t=1}^{n_i} ((X_i^t)^\T w^* + e_i^t) X_{ij}^t| > \lambda ) \\
		&=  \prob( |\frac{1}{n_i} \sum_{t=1}^{n_i} (\sum_{k=1}^d X_{ik}^t w^*_k + e_i^t) X_{ij}^t| > \lambda ) \\
		&= \prob( |\frac{1}{n_i} \sum_{t=1}^{n_i} ((X_{ij}^t)^2 w_j^* + \sum_{k \in S^*, k \ne j} X_{ik}^t X_{ij}^t w^*_k + e_i^t X_{ij}^t)| > \lambda ) \\
		&= \prob( |\frac{1}{n_i} \sum_{t=1}^{n_i} (X_{ij}^t)^2 w_j^* + \sum_{k \in S^*, k \ne j} \frac{1}{n_i} \sum_{t=1}^{n_i} X_{ik}^t X_{ij}^t w^*_k + \frac{1}{n_i} \sum_{t=1}^{n_i} e_i^t X_{ij}^t| > \lambda )
		\end{split}
		\end{align}
		Recall that in this setting the covariance matrix for $X_i$ is $\Sigma^i$ with diagonal entries $\Sigma^i_{jj} \equiv {\sigma^i_{jj}}^2, \forall j \in \seq{d}$ and non-diagonal entries $\Sigma^i_{jk} \equiv 0, \forall j,k \in \seq{d}, j \ne k$. Let $D_{ij} \triangleq w_j^* \sigma_{jj}^i$. Then,
		\begin{align}
		\begin{split}
		\prob( |\alphahat_{ij}| > \lambda ) &= \prob( |w_j^* \frac{1}{n_i} \sum_{t=1}^{n_i} ((X_{ij}^t)^2 - {\sigma^i_{jj}}^2)  + \sum_{k \in S^*, k \ne j} w^*_k \frac{1}{n_i} \sum_{t=1}^{n_i} X_{ik}^t X_{ij}^t + \frac{1}{n_i} \sum_{t=1}^{n_i} e_i^t X_{ij}^t + D_{ij}| > \lambda )
		\end{split}
		\end{align}
		Using the reverse triangle inequality $|a + b| \geq |a| - |b|$ recursively,
		\begin{align}
		\begin{split}	
		&\geq \prob( |D_{ij}| - |w_j^* \frac{1}{n_i} \sum_{t=1}^{n_i} ((X_{ij}^t)^2 - {\sigma^i_{jj}}^2)| - \sum_{k \in S^*, k \ne j} |w^*_k \frac{1}{n_i} \sum_{t=1}^{n_i} X_{ik}^t X_{ij}^t| - | \frac{1}{n_i} \sum_{t=1}^{n_i} e_i^t X_{ij}^t| > \lambda ) \\
		&\geq \prob( |D_{ij}| > \lambda + \delta_j + \delta_1 + \delta_e  \wedge |w_j^* \frac{1}{n_i} \sum_{t=1}^{n_i} ((X_{ij}^t)^2 - {\sigma^i_{jj}}^2)|<  \delta_j \wedge | \sum_{k \in S^*, k \ne j} w^*_k \frac{1}{n_i} \sum_{t=1}^{n_i} (X_{ik}^t X_{ij}^t )|\\ 
		&< \delta_1  \wedge | \frac{1}{n_i} \sum_{t=1}^{n_i} e_i^t X_{ij}^t| < \delta_e ) \\
		&\geq 1 - \prob(|D_{ij}| \leq \lambda + \delta_j + \delta_1 + \delta_e) - \prob(|w_j^* \frac{1}{n_i} \sum_{t=1}^{n_i} ((X_{ij}^t)^2 - {\sigma^i_{jj}}^2)|\geq  \delta_j   ) - \\
		& \prob( |\sum_{k \in S^*, k \ne j}  w^*_k \frac{1}{n_i} \sum_{t=1}^{n_i} X_{ik}^t X_{ij}^t | \geq \delta_1 ) - \prob(| \frac{1}{n_i} \sum_{t=1}^{n_i} e_i^t X_{ij}^t| \geq \delta_e ) 
		\end{split}
		\end{align} 
		
		The following Lemmas \ref{lem:bound x^2 term}, \ref{lem:uncorrelated wkXikXij} and \ref{lem:error term} provide concentration bounds for separate terms above.
		\begin{lemma}
			\label{lem:bound x^2 term}
			For $0 \leq \delta_j \leq 8 |w_j^*| \rho_i^2, \forall i \in \seq{g}, \forall j \in S^*$, 
			\begin{align}
			\begin{split}
			\prob(|w_j^* \frac{1}{n_i} \sum_{t=1}^{n_i} ((X_{ij}^t)^2 - {\sigma^i_{jj}}^2)|\geq \delta_j) \leq 2 \exp( -\frac{n_i(\frac{\delta_j}{|w_j^*|\rho_i^2} )^2}{64} ) 
			\end{split}
			\end{align}
		\end{lemma}
		
		We take $\delta_j = 8 |w_j^*| \rho_i^2 \delta$, then
		\begin{align}
		\begin{split}
		\prob(|\frac{1}{n_i} \sum_{t=1}^{n_i} ((\frac{X_{ij}^t}{\rho_i})^2 - (\frac{\sigma^i_{jj}}{\rho_i})^2)|\geq \frac{\delta_j}{|w_j^*|\rho_i^2} ) \leq 2\exp(-n_i\delta^2), \quad \forall 0 \leq \delta  \leq 1
		\end{split}
		\end{align}
		
		\begin{lemma}
			\label{lem:uncorrelated wkXikXij}
			For $0 \leq \delta_1 \leq 8 \rho_i^2 \sqrt{\sum_{k \in S^*, k \ne j} {w_k^*}^2}, \forall i \in \seq{g}, \forall j \in S^*$, if predictors are mutually independent of each other then
			\begin{align}
			\begin{split}
			\prob( |\sum_{k \in S^*, k \ne j} w^*_k \frac{1}{n_i} \sum_{t=1}^{n_i} X_{ik}^t X_{ij}^t | \geq \delta_1 ) \leq 2 \exp(- \frac{ n_i (\frac{\delta_1}{\rho_i^2 \sqrt{\sum_{k\in S^*, k \ne j} {w_k^*}^2} })^2 }{64})
			\end{split}
			\end{align}
		\end{lemma}
		
		We take $\delta_1 = 8 \delta \rho_i^2 \sqrt{\sum_{k\in S^*, k \ne j} {w_k^*}^2}$, which gives us
		\begin{align}
		\begin{split}
		\prob( |\sum_{k \in S^*, k \ne j} w^*_k \frac{1}{n_i} \sum_{t=1}^{n_i} X_{ik}^t X_{ij}^t | \geq \delta_1 ) \leq 2 \exp(-n_i \delta^2), \quad \forall 0 \leq \delta \leq 1
		\end{split}
		\end{align}
		
		\begin{lemma}
			\label{lem:error term}
			For $0 \leq \delta_e \leq 8 |\eta_i \rho_i|, \forall i \in \seq{g}, \forall j \in S^*$,
			\begin{align}
			\begin{split}
			\prob(| \frac{1}{n_i} \sum_{t=1}^{n_i} e_i^t X_{ij}^t| \geq  \delta_e ) \leq  2 \exp(- \frac{n_i (\frac{\delta_e}{|\eta_i\rho_i|})^2}{64})
			\end{split}
			\end{align}
		\end{lemma}
		
		We take $\delta_e = 8 |\eta_i \rho_i| \delta$, then
		\begin{align}
		\begin{split}
		\prob(| \frac{1}{n_i} \sum_{t=1}^{n_i} \frac{e_i^t}{\eta_i} \frac{X_{ij}^t}{\rho_i}| \geq \frac{\delta_e}{|\eta_i\rho_i|} ) \leq 2 \exp(- n_i\delta^2), \quad \forall 0 \leq \delta \leq 1
		\end{split}
		\end{align}
		
		By making the appropriate substitutions for $\delta_j, \delta_1$ and $\delta_e$ and noting that $|w_j^* {\sigma_{jj}^i}^2| > \lambda + 8 |w_j^*| \rho_i^2 \delta + 8  \rho_i^2 \sqrt{\sum_{k\in S^*, k \ne j} {w_k^*}^2} \delta + 8 |\eta_i\rho_i| \delta$, we can write:
		\begin{align}
		\begin{split}
		\E(R_j) &\geq 1 - \frac{6}{g} \sum_{i=1}^g \exp(-n_i\delta^2)
		\end{split}
		\end{align}
		
		It follows that if we take $n_i > \frac{1}{\delta^2} \log 12$ then $\E(R_j) > \frac{1}{2}$.
	\end{proof}

	\section{Proof of Lemma \ref{lem:bound x^2 term}}
	\label{sec:proof of lemma bound x^2 term}
	\emph{\paragraph{Lemma \ref{lem:bound x^2 term}}
		For $0 \leq \delta_j \leq 8 |w_j^*| \rho_i^2, \forall i \in \seq{g}, \forall j \in S^*$, 
		\begin{align}
		\begin{split}
		\prob(|w_j^* \frac{1}{n_i} \sum_{t=1}^{n_i} ((X_{ij}^t)^2 - {\sigma^i_{jj}}^2)|\geq \delta_j) \leq 2 \exp( -\frac{n_i(\frac{\delta_j}{|w_j^*|\rho_i^2} )^2}{64} ) 
		\end{split}
		\end{align}
	}
	\begin{proof}
		Observe that,
		\begin{align}
		\begin{split}
		\prob(|w_j^* \frac{1}{n_i} \sum_{t=1}^{n_i} ((X_{ij}^t)^2 - {\sigma^i_{jj}}^2)|\geq \delta_j   ) &= \prob(|\frac{1}{n_i} \sum_{t=1}^{n_i} ((X_{ij}^t)^2 - {\sigma^i_{jj}}^2)|\geq \frac{\delta_j}{|w_j^*|}   ) \\
		&= \prob(|\frac{1}{n_i} \sum_{t=1}^{n_i} ((\frac{X_{ij}^t}{\rho_i})^2 - (\frac{\sigma^i_{jj}}{\rho_i})^2)|\geq \frac{\delta_j}{|w_j^*|\rho_i^2}   )
		\end{split}
		\end{align}
		Note that $\frac{X_{ij}}{\rho_i}$ is a zero mean sub-Gaussian random variable with a variance proxy $1$. This implies that $(\frac{X_{ij}}{\rho_i})^2$ is a sub-exponential random variable with parameters $(4\sqrt{2}, 4)$. Thus, using concentration bounds for sub-exponential random variables\cite{wainwright2015chapter}, we can write:
		\begin{align}
		\begin{split}
		\prob(|\frac{1}{n_i} \sum_{t=1}^{n_i} ((\frac{X_{ij}^t}{\rho_i})^2 - (\frac{\sigma^i_{jj}}{\rho_i})^2)|\geq \frac{\delta_j}{|w_j^*|\rho_i^2} ) \leq 2\exp(-\frac{n_i(\frac{\delta_j}{|w_j^*|\rho_i^2} )^2}{64}), \quad \forall 0 \leq \frac{\delta_j}{|w_j^*|\rho_i^2}  \leq 8
		\end{split}
		\end{align}  
	\end{proof}
	
	\section{Proof of Lemma \ref{lem:uncorrelated wkXikXij}}
	\label{sec:proof of lem:uncorrelated wkXikXij}
	\emph{\paragraph{Lemma \ref{lem:uncorrelated wkXikXij}}
		For $0 \leq \delta_1 \leq 8 \rho_i^2 \sqrt{\sum_{k \in S^*, k \ne j} {w_k^*}^2}, \forall i \in \seq{g}, \forall j \in S^*$, if predictors are mutually independent of each other then
		\begin{align}
		\begin{split}
		\prob( |\sum_{k \in S^*, k \ne j} w^*_k \frac{1}{n_i} \sum_{t=1}^{n_i} X_{ik}^t X_{ij}^t | \geq \delta_1 ) \leq 2 \exp(- \frac{ n_i (\frac{\delta_1}{\rho_i^2 \sqrt{\sum_{k\in S^*, k \ne j} {w_k^*}^2} })^2 }{64})
		\end{split}
		\end{align}
	}
	\begin{proof}
		\label{proof:uncorrelated wkXikXij}
		Note that,
		\begin{align}
		\begin{split}
		\sum_{k\in S^*, k \ne j} w_k^* \frac{1}{n_i} \sum_{t=1}^{n_i} X_{ik}^t X_{ij}^t = \frac{1}{n_i} \sum_{t=1}^{n_i} (\sum_{k\in S^*, k\ne j} w_k^* X_{ik}^t) X_{ij}^t
		\end{split}
		\end{align}
		Here $X_{ik}^t$ is a zero mean sub-Gaussian random variable with variance proxy $\rho_i$ and $(\sum_{k\in S^*, k\ne j} w_k^* X_{ik}^t)$ is a zero mean sub-Gaussian random variable with variance proxy $\sqrt{\sum_{k\in S^*, k \ne j} {w_k^*}^2} \rho_i $. Thus,
		\begin{align}
		\begin{split}
		\prob( |\sum_{k \in S^*, k \ne j} w^*_k \frac{1}{n_i} \sum_{t=1}^{n_i} X_{ik}^t X_{ij}^t | \geq \delta_1 ) &= \prob( |\frac{1}{n_i} \sum_{t=1}^{n_i} (\sum_{k\in S^*, k\ne j} w_k^* X_{ik}^t) X_{ij}^t | \geq \delta_1 ) \\
		&= \prob( |\frac{1}{n_i} \sum_{t=1}^{n_i} \frac{(\sum_{k\in S^*, k\ne j} w_k^* X_{ik}^t)}{\sqrt{\sum_{k\in S^*, k \ne j} {w_k^*}^2} \rho_i} \frac{X_{ij}^t}{\rho_i} | \geq \frac{\delta_1}{\rho_i^2 \sqrt{\sum_{k\in S^*, k \ne j} {w_k^*}^2} } ) 
		\end{split}
		\end{align}
		where $\frac{(\sum_{k\in S^*, k\ne j} w_k^* X_{ik}^t)}{\sqrt{\sum_{k\in S^*, k \ne j} {w_k^*}^2} \rho_i}$ and $\frac{X_{ij}^t}{\rho_i}$ are zero mean sub-Gaussian random variables with unit variance proxy. Thus their product is sub-exponential random variable with parameters $(4\sqrt{2}, 4)$. Using concentration bound for sub-exponential random variables \cite{wainwright2015chapter}, we can write:
		\begin{align}
		\begin{split}
		&\prob( |\frac{1}{n_i} \sum_{t=1}^{n_i} \frac{(\sum_{k\in S^*, k\ne j} w_k^* X_{ik}^t)}{\sqrt{\sum_{k\in S^*, k \ne j} {w_k^*}^2} \rho_i} \frac{X_{ij}^t}{\rho_i} | \geq \frac{\delta_1}{\rho_i^2 \sqrt{\sum_{k\in S^*, k \ne j} {w_k^*}^2} } )  \leq 2 \exp(- \frac{ n_i (\frac{\delta_1}{\rho_i^2 \sqrt{\sum_{k\in S^*, k \ne j} {w_k^*}^2} })^2 }{64}) \\
		&\forall 0 \leq \frac{\delta_1}{\rho_i^2 \sqrt{\sum_{k\in S^*, k \ne j} {w_k^*}^2} } \leq 8
		\end{split}
		\end{align} 
	\end{proof}

	\section{Proof of Lemma \ref{lem:error term}}
	\label{sec:proof of lemma error term}
	\emph{\paragraph{Lemma \ref{lem:error term}}
		For $0 \leq \delta_e \leq 8 |\eta_i \rho_i|, \forall i \in \seq{g}, \forall j \in S^*$,
		\begin{align}
		\begin{split}
		\prob(| \frac{1}{n_i} \sum_{t=1}^{n_i} e_i^t X_{ij}^t| \geq  \delta_e ) \leq  2 \exp(- \frac{n_i (\frac{\delta_e}{|\eta_i\rho_i|})^2}{64})
		\end{split}
		\end{align}
	}
	\begin{proof}
		\label{proof:error term}
		Note that,
		\begin{align}
		\begin{split}
		\prob(| \frac{1}{n_i} \sum_{t=1}^{n_i} e_i^t X_{ij}^t| \geq  \delta_e ) &= \prob(| \frac{1}{n_i} \sum_{t=1}^{n_i} \frac{e_i^t}{\eta_i} \frac{X_{ij}^t}{\rho_i}| \geq \frac{\delta_e}{|\eta_i\rho_i|} )
		\end{split}
		\end{align}
		Again $\frac{X_{ij}}{\rho_i}$ and $\frac{e_i}{\eta_i}$ are mutually independent zero mean sub-Gaussian random variables with a variance proxy $1$. Thus, $\frac{e_i^t}{\eta_i} \frac{X_{ij}^t}{\rho_i}$ is a sub-exponential random variable with parameters $(4\sqrt{2}, 4)$. Thus, using concentration bounds for sub-exponential random variables\cite{wainwright2015chapter}, we can write:
		\begin{align}
		\begin{split}
		\prob(| \frac{1}{n_i} \sum_{t=1}^{n_i} \frac{e_i^t}{\eta_i} \frac{X_{ij}^t}{\rho_i}| \geq \frac{\delta_e}{|\eta_i\rho_i|} ) \leq 2 \exp(- \frac{n_i (\frac{\delta_e}{|\eta_i\rho_i|})^2}{64}), \quad \forall 0 \leq \frac{\delta_e}{|\eta_i\rho_i|} \leq 8
		\end{split}
		\end{align}
	\end{proof}
	
	\section{Proof of Lemma \ref{lem:nonsupport mcdiarmid}}
	\label{sec:proof of lemma nonsupport mcdiarmid}
	\emph{\paragraph{Lemma \ref{lem:nonsupport mcdiarmid}}
		For $j \in S^*_c$, let $\E(R_j) < \frac{1}{2}$, then $R_j \leq \frac{1}{2}$ with probability at least $1 - 2 \exp( - 2g (\frac{1}{2} - \E(R_j))^2)$.
	}
	\begin{proof}
		\label{proof:nonsupport mcdiarmid}
		We again note that,
		\begin{align}
		\begin{split}
		\prob(R_j \geq \frac{1}{2}) &= \prob(R_j - \E(R_j) \geq \frac{1}{2} - \E(R_j)) \\
		&\leq \prob(|R_j - \E(R_j)| \geq \frac{1}{2} - \E(R_j))
		\end{split}
		\end{align}
		The last inequality holds as long as $\E(R_j) < \frac{1}{2}$. Again, by noting that for a fixed $w^*$, $R_j = f(R_{1j}, R_{2j}, \cdots, R_{gj})$ such that 
		\begin{align}
		\begin{split}
		(\forall k \in \seq{g}),\quad \sup_{R_{1j}, \cdots, R_{kj}, R_{kj}',\cdots, R_{gj}}|f(R_{1j}, \cdots, R_{kj},\cdots, R_{gj}) - f(R_{1j}, \cdots, R_{kj}',\cdots, R_{gj})| \leq \frac{1}{g}
		\end{split}
		\end{align}
		and using McDiarmid's inequality \cite{mcdiarmid1989method}, we can write the following:
		\begin{align}
		\begin{split}
		\prob(|R_j - \E(R_j)| \geq \frac{1}{2} - \E(R_j)) \leq 2 \exp( - 2g (\frac{1}{2} - \E(R_j))^2)
		\end{split}
		\end{align}
	\end{proof}

	\section{Proof of Lemma \ref{lem:nonsupport bound E(Rj)}}
	\label{sec:proof of lemma nonsupport bound E(Rj)}
	\emph{\paragraph{Lemma \ref{lem:nonsupport bound E(Rj)}}
		For any $j \in S^*_c$ and $0 < \delta \leq 1$, if predictors are mutually independent of each other and if
		\begin{align}
		\begin{split}
		\lambda > 8 \delta \rho_i^2 \sqrt{\sum_{k\in S^*} w_k^2} + 8 |\eta_i\rho_i|\delta
		\end{split}
		\end{align}  
		then $\E(R_j) \leq \frac{4}{g} \sum_{i=1}^g \exp(-n_i \delta^2)$. Furthermore, for $n_i = \calO(\frac{1}{\delta^2})$, we have $\E(R_j) < \frac{1}{2}$.
	}
	\begin{proof}
		\label{proof:nonsupport bound E(Rj)}
		Like the proof of Lemma \ref{lem:support bound on E(Rj)}, we have the same formula for $\E(R_j)$ with the difference that $j \in S^*_c$, i.e.,
		\begin{align}
		\begin{split}
		\E(R_j) = \frac{1}{g} \sum_{i=1}^g \prob( |\alphahat_{ij}| > \lambda )
		\end{split}
		\end{align} 
		This time, we will put an upper bound on $\prob( |\alphahat_{ij}| > \lambda )$.
		\begin{align}
		\begin{split}
		\prob( |\alphahat_{ij}| > \lambda ) &= \prob( |\frac{1}{n_i} \sum_{t=1}^{n_i} y_i^t X_{ij}^t| > \lambda ) \\
		& \text{Expanding $y_i^t$ using equation \eqref{eq:generative model}} \\
		&= \prob( |\frac{1}{n_i} \sum_{t=1}^{n_i} ((X_i^t)^\T w^* + e_i^t) X_{ij}^t| > \lambda ) \\
		&=  \prob( |\frac{1}{n_i} \sum_{t=1}^{n_i} (\sum_{k=1}^d X_{ik}^t w^*_k + e_i^t) X_{ij}^t| > \lambda ) \\
		&= \prob( |\frac{1}{n_i}(\sum_{k \in S^*} X_{ik}^t X_{ij}^t w^*_k + e_i^t X_{ij}^t)| > \lambda ) \\
		&= \prob( | \sum_{k \in S^*} \frac{1}{n_i} \sum_{t=1}^{n_i} X_{ik}^t X_{ij}^t w^*_k + \frac{1}{n_i} \sum_{t=1}^{n_i} e_i^t X_{ij}^t| > \lambda )
		\end{split}
		\end{align}
		Using the triangle inequality $|a+b| \leq |a| + |b|$, we can rewrite the above equation as:
		\begin{align}
		\begin{split}
		\prob( |\alphahat_{ij}| > \lambda ) &\leq \prob( |\sum_{k \in S^*} w^*_k \frac{1}{n_i} \sum_{t=1}^{n_i} X_{ik}^t X_{ij}^t |   + |\frac{1}{n_i} \sum_{t=1}^{n_i} e_i^t X_{ij}^t|  > \lambda ) \\
		&= 1 - \prob( | \sum_{k \in S^*} w^*_k \frac{1}{n_i} \sum_{t=1}^{n_i} X_{ik}^t X_{ij}^t |   + |\frac{1}{n_i} \sum_{t=1}^{n_i} e_i^t X_{ij}^t|  \leq \lambda ) \\
		&\leq 1 - \prob( | \sum_{k \in S^*} w^*_k \frac{1}{n_i} \sum_{t=1}^{n_i} X_{ik}^t X_{ij}^t | \leq \delta_1  \wedge |\frac{1}{n_i} \sum_{t=1}^{n_i} e_i^t X_{ij}^t| \leq \delta_e \wedge 0 \leq \lambda - \delta_1 - \delta_e ) \\
		&\leq  \prob(|\sum_{k \in S^*} w^*_k \frac{1}{n_i} \sum_{t=1}^{n_i} X_{ik}^t X_{ij}^t | \geq \delta_1 ) + \prob(|\frac{1}{n_i} \sum_{t=1}^{n_i} e_i^t X_{ij}^t| \geq \delta_e) + \prob(0 \geq \lambda - \delta_1 - \delta_e)
		\end{split}
		\end{align}
		
		The following Lemmas \ref{lem:nonsupport uncorrelated wkXikXij} and \ref{lem:nonsupport error term} provide concentration bounds for the separate terms above.
		\begin{lemma}
			\label{lem:nonsupport uncorrelated wkXikXij}
			For $0 \leq \delta_1 \leq 8 \rho_i^2 \sqrt{\sum_{k \in S^*} {w_k^*}^2}, \forall i \in \seq{g}, \forall j \in S^*_c$, if predictors are mutually independent of each other then
			\begin{align}
			\begin{split}
			\prob( |\sum_{k \in S^*} w^*_k \frac{1}{n_i} \sum_{t=1}^{n_i} X_{ik}^t X_{ij}^t | \geq \delta_1 ) \leq 2 \exp(- \frac{ n_i (\frac{\delta_1}{\rho_i^2 \sqrt{\sum_{k\in S^*} {w_k^*}^2} })^2 }{64})
			\end{split}
			\end{align}
		\end{lemma}
		
		We take $\delta_1 = 8 \rho_i^2 \sqrt{\sum_{k \in S^*} {w_k^*}^2} \delta$ for $0 < \delta < 1$. Then,
		\begin{align}
		\begin{split}
		\prob( |\sum_{k \in S^*} w^*_k \frac{1}{n_i} \sum_{t=1}^{n_i} X_{ik}^t X_{ij}^t | \geq \delta_1 ) \leq 2 \exp(-n_i \delta^2)
		\end{split}
		\end{align}
		
		\begin{lemma}
			\label{lem:nonsupport error term}
			For $0 \leq \delta_e \leq 8 |\eta_i \rho_i|, \forall i \in \seq{g}, \forall j \in S^*_c$,
			\begin{align}
			\begin{split}
			\prob(| \frac{1}{n_i} \sum_{t=1}^{n_i} e_i^t X_{ij}^t| \geq  \delta_e ) \leq  2 \exp(- \frac{n_i (\frac{\delta_e}{|\eta_i\rho_i|})^2}{64})
			\end{split}
			\end{align}
		\end{lemma}
		
		We take $\delta_e = 8 |\eta_i \rho_i| \delta$ for $0 < \delta < 1$, then
		\begin{align}
		\begin{split}
		\prob(| \frac{1}{n_i} \sum_{t=1}^{n_i} e_i^t X_{ij}^t| \geq  \delta_e ) \leq  2 \exp(-n_i \delta^2)
		\end{split}
		\end{align}
		
		Using results of Lemmas \ref{lem:nonsupport uncorrelated wkXikXij} and \ref{lem:nonsupport error term} and making the appropriate substitutions for $\delta_1$ and $\delta_e$ and noticing that $ \lambda > 8 \delta \rho_i^2 \sqrt{\sum_{k\in S^*} w_k^2} + 8 |\eta_i\rho_i|\delta$, we can write
		\begin{align}
		\begin{split}
		\E(R_j) \leq \frac{4}{g} \sum_{i=1}^g \exp(-n_i\delta^2) 
		\end{split}
		\end{align}   
		Furthermore, if we have $n_i > \frac{1}{\delta^2} \log 8$ then $\E(R_j) < \frac{1}{2}$.
	\end{proof}

	\section{Proof of Lemma \ref{lem:nonsupport uncorrelated wkXikXij}}
	\label{sec:proof of lemma nonsupport uncorrelated wkXikXij}
	\emph{\paragraph{Lemma \ref{lem:nonsupport uncorrelated wkXikXij}}
		For $0 \leq \delta_1 \leq 8 \rho_i^2 \sqrt{\sum_{k \in S^*} {w_k^*}^2}, \forall i \in \seq{g}, \forall j \in S^*_c$, if predictors are mutually independent of each other then
		\begin{align}
		\begin{split}
		\prob( |\sum_{k \in S^*} w^*_k \frac{1}{n_i} \sum_{t=1}^{n_i} X_{ik}^t X_{ij}^t | \geq \delta_1 ) \leq 2 \exp(- \frac{ n_i (\frac{\delta_1}{\rho_i^2 \sqrt{\sum_{k\in S^*} {w_k^*}^2} })^2 }{64})
		\end{split}
		\end{align}
	}
	\begin{proof}
		We follow the same procedure as the proof of Lemma \ref{lem:uncorrelated wkXikXij}.
	\end{proof}

	\section{Proof of Lemma \ref{lem:nonsupport error term}}
	\label{sec:proof of lemma nonsupport error term}
	\emph{\paragraph{Lemma \ref{lem:nonsupport error term}}
		For $0 \leq \delta_e \leq 8 |\eta_i \rho_i|, \forall i \in \seq{g}, \forall j \in S^*_c$,
		\begin{align}
		\begin{split}
		\prob(| \frac{1}{n_i} \sum_{t=1}^{n_i} e_i^t X_{ij}^t| \geq  \delta_e ) \leq  2 \exp(- \frac{n_i (\frac{\delta_e}{|\eta_i\rho_i|})^2}{64})
		\end{split}
		\end{align}
	}
	\begin{proof}
		We follow the same procedure as the proof of Lemma \ref{lem:error term}.
	\end{proof}
	
	\section{Proof of Lemma \ref{lem:support bound on E(Rj) correlated}}
	\label{sec:proof of lemma support bound on E(Rj) correlated}
	\emph{\paragraph{Lemma \ref{lem:support bound on E(Rj) correlated}}
		For any $j \in S^*$ and some $0 < \delta \leq \frac{1}{\sqrt{2}}$, if $\forall j \in S^*$,
		\begin{align}
		\begin{split}
		&0 < \lambda < |(w^*_j{\sigma^i_{jj}}^2 + \sum_{k\in S^*, k\ne j} w^*_k \sigma^i_{jk})| - 8 |w_j^*| \rho_i^2 \delta - \sum_{k\in S^*, k\ne j} 8 \sqrt{2} |w_k^*| (1 + 4 \max_j \frac{\rho_i^2}{{\sigma^i_{jj}}^2}) \\
		&\max_j {\sigma^i_{jj}}^2 \delta - 8 |\eta_i \rho_i| \delta
		\end{split}
		\end{align} 
		then $ \E(R_j) \geq 1 - \frac{4s}{g} \sum_{i=1}^g \exp(-n_i\delta^2 ) $. Furthermore, for $n_i = \calO(\frac{1}{\delta^2}\log s)$, we have $\E(R_j) > \frac{1}{2}$.
	}
	\begin{proof}
		\label{proof:support bound on E(Rj) correlated}
		We know from the proof of Lemma \ref{lem:support bound on E(Rj)} that
		\begin{align}
		\begin{split}
		\E(R_j) = \frac{1}{g} \sum_{i=1}^g \prob(|\alphahat_{ij}| > \lambda)
		\end{split}
		\end{align}
		and
		\begin{align}
		\begin{split}
		\prob( |\alphahat_{ij}| > \lambda ) &= \prob( |\frac{1}{n_i} \sum_{t=1}^{n_i} (X_{ij}^t)^2 w_j^* + \sum_{k \in S^*, k \ne j} \frac{1}{n_i} \sum_{t=1}^{n_i} X_{ik}^t X_{ij}^t w^*_k + \frac{1}{n_i} \sum_{t=1}^{n_i} e_i^t X_{ij}^t| > \lambda )
		\end{split}
		\end{align}
		Let $D_{ij} = w^*_j\E(X_{ij}^2) + \sum_{k\in S^*, k\ne j} w^*_k \E(X_{ik}X_{ij}) =  w^*_j{\sigma^i_{jj}}^2 + \sum_{k\in S^*, k\ne j} w^*_k \sigma^i_{jk}$. Then,
		\begin{align}
		\begin{split}
		\prob( |\alphahat_{ij}| > \lambda ) &= \prob( |w_j^* \frac{1}{n_i} \sum_{t=1}^{n_i} ((X_{ij}^t)^2 - {\sigma^i_{jj}}^2)  + \sum_{k \in S^*, k \ne j} w^*_k \frac{1}{n_i} \sum_{t=1}^{n_i} (X_{ik}^t X_{ij}^t - \sigma^i_{jk} )\\
		&+ \frac{1}{n_i} \sum_{t=1}^{n_i} e_i^t X_{ij}^t + D_{ij}| > \lambda )
		\end{split}
		\end{align}
		Using the reverse triangle inequality $|a + b| \geq |a| - |b|$ recursively,
		\begin{align}
		\begin{split}	
		&\geq \prob( |D_{ij}| - |w_j^* \frac{1}{n_i} \sum_{t=1}^{n_i} ((X_{ij}^t)^2 - {\sigma^i_{jj}}^2)| - \sum_{k \in S^*, k \ne j} |w^*_k \frac{1}{n_i} \sum_{t=1}^{n_i} (X_{ik}^t X_{ij}^t - \sigma^i_{jk} )| - | \frac{1}{n_i} \sum_{t=1}^{n_i} e_i^t X_{ij}^t| > \lambda ) \\
		&\geq \prob( |D_{ij}| > \lambda + \delta_j + \sum_{k \in S^*, k \ne j} \delta_k + \delta_e  \wedge |w_j^* \frac{1}{n_i} \sum_{t=1}^{n_i} ((X_{ij}^t)^2 - {\sigma^i_{jj}}^2)|<  \delta_j \wedge (\forall k \in S^*, k \ne j)\\
		& |w^*_k \frac{1}{n_i} \sum_{t=1}^{n_i} (X_{ik}^t X_{ij}^t - \sigma^i_{jk} )|< \delta_k  \wedge | \frac{1}{n_i} \sum_{t=1}^{n_i} e_i^t X_{ij}^t| < \delta_e ) \\
		&\geq 1 - \prob(|D_{ij}| \leq \lambda + \delta_j + \sum_{k \in S^*, k \ne j} \delta_k + \delta_e) - \prob(|w_j^* \frac{1}{n_i} \sum_{t=1}^{n_i} ((X_{ij}^t)^2 - {\sigma^i_{jj}}^2)|\geq  \delta_j   ) - \\
		& \sum_{k \in S^*, k \ne j} \prob( |w^*_k \frac{1}{n_i} \sum_{t=1}^{n_i} (X_{ik}^t X_{ij}^t - \sigma^i_{jk} )| \geq \delta_k ) - \prob(| \frac{1}{n_i} \sum_{t=1}^{n_i} e_i^t X_{ij}^t| \geq \delta_e ) 
		\end{split}
		\end{align}
		Now we focus on concentration bounds for individual terms. Note that we already have concentration bounds for $\prob(|w_j^* \frac{1}{n_i} \sum_{t=1}^{n_i} ((X_{ij}^t)^2 - {\sigma^i_{jj}}^2)|\geq  \delta_j   ) $ and $\prob(| \frac{1}{n_i} \sum_{t=1}^{n_i} e_i^t X_{ij}^t| \geq \delta_e ) $ from Lemmas \ref{lem:bound x^2 term} and \ref{lem:error term} respectively. The following lemma provides concentration bound for $\prob( |w^*_k \frac{1}{n_i} \sum_{t=1}^{n_i} (X_{ik}^t X_{ij}^t - \sigma^i_{jk} )| \geq  \delta_k  )$. 
		\begin{lemma}
			\label{lem:support correlated wkXikXij}
			For $0 \leq \delta_k \leq 8 \sqrt{2} |w_k^*| (1 + 4 \max_j \frac{\rho_i^2}{{\sigma^i_{jj}}^2}) \max_j {\sigma^i_{jj}}^2, \forall i \in \seq{g}, \forall j \in S^*$, then
			\begin{align}
			\begin{split}
			\prob( |w^*_k \frac{1}{n_i} \sum_{t=1}^{n_i} (X_{ik}^t X_{ij}^t - \sigma^i_{jk} )| \geq  \delta_k  ) \leq 4 \exp( - \frac{n_i (\frac{\delta_k}{|w^*_k|})^2}{128(1 + 4 \max_j \frac{\rho_i^2}{{\sigma^i_{jj}}^2})^2 \max_j{\sigma^i_{jj}}^4})
			\end{split}
			\end{align}
		\end{lemma}
		
		We take $\delta_k = 8 \sqrt{2} |w_k^*| (1 + 4 \max_j \frac{\rho_i^2}{{\sigma^i_{jj}}^2}) \max_j {\sigma^i_{jj}}^2 \delta$, then
		\begin{align}
		\begin{split}
		&\prob( |\frac{1}{n_i} \sum_{t=1}^{n_i} (X_{ik}^t X_{ij}^t - \sigma^i_{jk} )| \geq \frac{\delta_k}{|w^*_k|} )  \leq 4 \exp( - n_i \delta^2), \forall 0 \leq  \delta \leq \frac{1}{\sqrt{2}}
		\end{split}
		\end{align}
		Also taking $\delta_j =8 |w_j^*| \rho_i^2 \delta$ and $\delta_e = 8 |\eta_i \rho_i| \delta $ in Lemmas \ref{lem:bound x^2 term} and \ref{lem:error term} respectively and noting that
		\begin{align}
		\begin{split}
		&0 < \lambda < |(w^*_j{\sigma^i_{jj}}^2 + \sum_{k\in S^*, k\ne j} w^*_k \sigma^i_{jk})| - 8 |w_j^*| \rho_i^2 \delta - \sum_{k\in S^*, k\ne j} 8 \sqrt{2} |w_k^*| (1 + 4 \max_j \frac{\rho_i^2}{{\sigma^i_{jj}}^2}) \\
		&\max_j {\sigma^i_{jj}}^2 \delta - 8 |\eta_i \rho_i| \delta \;,
		\end{split}
		\end{align}
		we can write:
		\begin{align}
		\begin{split}
		\E(R_j) \geq 1 - \frac{4s}{g} \sum_{i=1}^g \exp(-n_i\delta^2 )
		\end{split}
		\end{align}
		It follows that if we take $n_i >  \frac{1}{\delta^2}\log 8s$, we get $\E(R_j) > \frac{1}{2}$.
	\end{proof}

	\section{Proof of Lemma \ref{lem:support correlated wkXikXij}}
	\label{sec:proof of lemma support correlated wkXikXij}
	\emph{\paragraph{Lemma \ref{lem:support correlated wkXikXij}}
		For $0 \leq \delta_k \leq 8 \sqrt{2} |w_k^*| (1 + 4 \max_j \frac{\rho_i^2}{{\sigma^i_{jj}}^2}) \max_j {\sigma^i_{jj}}^2, \forall i \in \seq{g}, \forall j \in S^*$, then
		\begin{align}
		\begin{split}
		\prob( |w^*_k \frac{1}{n_i} \sum_{t=1}^{n_i} (X_{ik}^t X_{ij}^t - \sigma^i_{jk} )| \geq  \delta_k  ) \leq 4 \exp( - \frac{n_i (\frac{\delta_k}{|w^*_k|})^2}{128(1 + 4 \max_j \frac{\rho_i^2}{{\sigma^i_{jj}}^2})^2 \max_j{\sigma^i_{jj}}^4})
		\end{split}
		\end{align}
	}
	\begin{proof}
		\label{proof:support correlated wkXikXij}
		Note that,
		\begin{align}
		\begin{split}
		\prob( |w^*_k \frac{1}{n_i} \sum_{t=1}^{n_i} (X_{ik}^t X_{ij}^t - \sigma^i_{jk} )| \geq  \delta_k  ) &= \prob( |\frac{1}{n_i} \sum_{t=1}^{n_i} (X_{ik}^t X_{ij}^t - \sigma^i_{jk} )| \geq \frac{\delta_k}{|w^*_k|} ) 
		\end{split}
		\end{align}
		Using Lemma 1 from \cite{ravikumar2011high}, we can write
		\begin{align}
		\begin{split}
		&\prob( |\frac{1}{n_i} \sum_{t=1}^{n_i} (X_{ik}^t X_{ij}^t - \sigma^i_{jk} )| \geq \frac{\delta_k}{|w^*_k|} )  \leq 4 \exp( - \frac{n_i (\frac{\delta_k}{|w^*_k|})^2}{128(1 + 4 \max_j \frac{\rho_i^2}{{\sigma^i_{jj}}^2})^2 \max_j{\sigma^i_{jj}}^4}), \\
		& \forall 0 \leq  \frac{\delta_k}{|w^*_k|} \leq 8\max_j {\sigma^i_{jj}}^2(1 + 4 \max_j \frac{\rho_i^2}{{\sigma^i_{jj}}^2})
		\end{split}
		\end{align}
	\end{proof}
	
	\section{Proof of Lemma \ref{lem:nonsupport bound E(Rj) correlated}}
	\label{sec:proof of lemma nonsupport bound E(Rj) correlated}
	\emph{\paragraph{Lemma \ref{lem:nonsupport bound E(Rj) correlated}}
		For any $j \in S^*_c$ and some $0 < \delta \leq \frac{1}{\sqrt{2}}$, if
		\begin{align}
		\begin{split}
		\lambda > |\sum_{k\in S^*} w^*_k \sigma^i_{jk}| + \sum_{k\in S^*} 8 \sqrt{2} |w_k^*| (1 + 4 \max_j \frac{\rho_i^2}{{\sigma^i_{jj}}^2}) \max_j {\sigma^i_{jj}}^2 \delta + 8 |\eta_i \rho_i| \delta
		\end{split}
		\end{align}  
		then $\E(R_j) \leq \frac{4s + 2}{g} \sum_{i=1}^g \exp(-n_i \delta^2)$. Furthermore, for $n_i = \calO(\frac{1}{\delta^2} \log s)$, we have $\E(R_j) < \frac{1}{2}$.
	}
	\begin{proof}
		\label{proof:nonsupport bound E(Rj) correlated}
		Like before,
		\begin{align}
		\begin{split}
		\E(R_j) = \frac{1}{g} \sum_{i=1}^g \prob( |\alphahat_{ij}| > \lambda )
		\end{split}
		\end{align} 
		and
		\begin{align}
		\begin{split}
		\prob( |\alphahat_{ij}| > \lambda ) = \prob( | \sum_{k \in S^*} \frac{1}{n_i} \sum_{t=1}^{n_i} X_{ik}^t X_{ij}^t w^*_k + \frac{1}{n_i} \sum_{t=1}^{n_i} e_i^t X_{ij}^t| > \lambda )
		\end{split}
		\end{align}
		Let $D_{ij} =  \sum_{k\in S^*} w^*_k \E(X_{ik}X_{ij}) =  \sum_{k\in S^*} w^*_k \sigma^i_{jk}$. Then,
		\begin{align}
		\begin{split}
		\prob( |\alphahat_{ij}| > \lambda ) &= \prob( |\sum_{k \in S^*} w^*_k \frac{1}{n_i} \sum_{t=1}^{n_i} (X_{ik}^t X_{ij}^t - \sigma^i_{jk} )   + \frac{1}{n_i} \sum_{t=1}^{n_i} e_i^t X_{ij}^t + D_{ij}| > \lambda ) 
		\end{split}
		\end{align}
		Using the triangle inequality $|a+b| \leq |a| + |b|$,
		\begin{align}
		\begin{split}
		\prob( |\alphahat_{ij}| > \lambda ) &\leq \prob( \sum_{k \in S^*} |w^*_k \frac{1}{n_i} \sum_{t=1}^{n_i} (X_{ik}^t X_{ij}^t - \sigma^i_{jk} )|   + |\frac{1}{n_i} \sum_{t=1}^{n_i} e_i^t X_{ij}^t| + |D_{ij}| > \lambda ) \\
		&= 1 - \prob( \sum_{k \in S^*} |w^*_k \frac{1}{n_i} \sum_{t=1}^{n_i} (X_{ik}^t X_{ij}^t - \sigma^i_{jk} )|   + |\frac{1}{n_i} \sum_{t=1}^{n_i} e_i^t X_{ij}^t| + |D_{ij}| \leq \lambda ) \\
		&\leq 1 - \prob( (\forall k \in S^*) |w^*_k \frac{1}{n_i} \sum_{t=1}^{n_i} (X_{ik}^t X_{ij}^t - \sigma^i_{jk} )| \leq \delta_k  \wedge |\frac{1}{n_i} \sum_{t=1}^{n_i} e_i^t X_{ij}^t| \leq \delta_e \wedge \\
		& |D_{ij}| \leq \lambda - \sum_{k \in S^*} \delta_k - \delta_e ) \\
		&\leq \sum_{k \in S^*} \prob(|w^*_k \frac{1}{n_i} \sum_{t=1}^{n_i} (X_{ik}^t X_{ij}^t - \sigma^i_{jk} )| \geq \delta_k ) + \prob(|\frac{1}{n_i} \sum_{t=1}^{n_i} e_i^t X_{ij}^t| \geq \delta_e) + \\
		& \prob(|D_{ij}| \geq \lambda - \sum_{k \in S^*} \delta_k - \delta_e)
		\end{split}
		\end{align}
		Observe that for entries in the non-support, we already have a concentration bound on $\prob(|\frac{1}{n_i} \sum_{t=1}^{n_i} e_i^t X_{ij}^t| \geq \delta_e)$ from Lemma \ref{lem:nonsupport error term}. The following lemma provides concentration bound on $\prob(|w^*_k \frac{1}{n_i} \sum_{t=1}^{n_i} (X_{ik}^t X_{ij}^t - \sigma^i_{jk} )| \geq \delta_k )$.
		\begin{lemma}
			\label{lem:nonsupport correlated wkXikXij}
			For $0 \leq \delta_k \leq 8 \sqrt{2} |w_k^*| (1 + 4 \max_j \frac{\rho_i^2}{{\sigma^i_{jj}}^2}) \max_j {\sigma^i_{jj}}^2, \forall i \in \seq{g}, \forall j \in S^*_c$, then
			\begin{align}
			\begin{split}
			\prob( |w^*_k \frac{1}{n_i} \sum_{t=1}^{n_i} (X_{ik}^t X_{ij}^t - \sigma^i_{jk} )| \geq  \delta_k  ) \leq 4 \exp( - \frac{n_i (\frac{\delta_k}{|w^*_k|})^2}{128(1 + 4 \max_j \frac{\rho_i^2}{{\sigma^i_{jj}}^2})^2 \max_j{\sigma^i_{jj}}^4})
			\end{split}
			\end{align}
		\end{lemma}
		
		We take $\delta_k = 8 \sqrt{2} |w_k^*| (1 + 4 \max_j \frac{\rho_i^2}{{\sigma^i_{jj}}^2}) \max_j {\sigma^i_{jj}}^2 \delta$, then
		\begin{align}
		\begin{split}
		&\prob( |\frac{1}{n_i} \sum_{t=1}^{n_i} (X_{ik}^t X_{ij}^t - \sigma^i_{jk} )| \geq \frac{\delta_k}{|w^*_k|} )  \leq 4 \exp( - n_i\delta^2), \forall 0 \leq  \delta \leq \frac{1}{\sqrt{2}}
		\end{split}
		\end{align}
		Taking $\delta_e = 8 |\eta_i \rho_i| \delta$ and noticing that
		\begin{align}
		\begin{split}
		\lambda > |\sum_{k\in S^*} w^*_k \sigma^i_{jk}| + \sum_{k\in S^*} 8 \sqrt{2} |w_k^*| (1 + 4 \max_j \frac{\rho_i^2}{{\sigma^i_{jj}}^2}) \max_j {\sigma^i_{jj}}^2 \delta + 8 |\eta_i \rho_i| \delta \;,
		\end{split}
		\end{align}
		we can write
		\begin{align}
		\begin{split}
		\E(R_j) \leq &\frac{4s + 2}{g} \sum_{i=1}^g ( \exp(-n_i \delta^2) ) 
		\end{split}
		\end{align}
		It also follows that if we have $n_i > \frac{1}{\delta^2}\log(8s + 4) $ then $\E(R_j) < \frac{1}{2}$.
	\end{proof}

	\section{Proof of Lemma \ref{lem:nonsupport correlated wkXikXij}}
	\label{sec:proof of lemma nonsupport correlated wkXikXij}
	\emph{\paragraph{Lemma \ref{lem:nonsupport correlated wkXikXij}}
		For $0 \leq \delta_k \leq 8 \sqrt{2} |w_k^*| (1 + 4 \max_j \frac{\rho_i^2}{{\sigma^i_{jj}}^2}) \max_j {\sigma^i_{jj}}^2, \forall i \in \seq{g}, \forall j \in S^*_c$, then
		\begin{align}
		\begin{split}
		\prob( |w^*_k \frac{1}{n_i} \sum_{t=1}^{n_i} (X_{ik}^t X_{ij}^t - \sigma^i_{jk} )| \geq  \delta_k  ) \leq 4 \exp( - \frac{n_i (\frac{\delta_k}{|w^*_k|})^2}{128(1 + 4 \max_j \frac{\rho_i^2}{{\sigma^i_{jj}}^2})^2 \max_j{\sigma^i_{jj}}^4})
		\end{split}
		\end{align}
	}
	\begin{proof}
		\label{proof:nonsupport correlated wkXikXij}
		The proof follows the same procedure as the proof of Lemma \ref{lem:support correlated wkXikXij}.
	\end{proof}
	
	\section{Real World Experiment}
	\label{sec:real world application}
	
	In this section, we demonstrate the effectiveness of our method to determine the support of a sparse linear regression setup in a real world data set. We used the BlogFeedback data set\footnote{Buza, K. (2014). Feedback Prediction for Blogs. In Data Analysis, Machine Learning and Knowledge Discovery (pp. 145-152). Springer International Publishing.} from \url{https://archive.ics.uci.edu/ml/datasets/BlogFeedback}. This data set contains features extracted from blog posts and the task is to predict how many comments the post will receive using these features. The details about the data set are as follows: 
	\begin{itemize}
		\item Number of training samples: $52397$
		\item Number of features (after removing all zeros columns): $276$
	\end{itemize}
	
	\paragraph{Constructing a True Support.} Since the true support for the parameter vector is unknown for real world data, we constructed a ``centralized'' support for comparison by running LASSO on the complete training data set. This simulates a centralized server with access to all the data. The ``centralized'' support contains $40$ elements. Our goal is to recover this ``centralized'' support by running our method in a federated setting.
	
	\paragraph{Performance Measures.} We measure the performance of our method by measuring recall, precision and $F_1$-score with respect to our recovered ``federated'' support. These performance measures are defined as follows:
	\begin{align}
	\begin{split}
	\text{Recall} &= \frac{\text{Number of elements in the ``federated'' support that are in the ``centralized'' support}}{\text{Number of elements in the ``centralized'' support}} \\
	\text{Precision} &= \frac{\text{Number of elements in the ``federated'' support that are in the ``centralized'' support}}{\text{Number of elements in the ``federated'' support}} \\
	F_1\text{-score} &= \frac{2 \times \text{Recall} \times \text{Precision}}{\text{Recall} + \text{Precision}} \\
	\end{split}
	\end{align}
	
	\paragraph{Case 1.} For the first experiment, we divided the dataset randomly into $523$ clients with each client containing $100$ samples (except the last one which contains more to account for imbalance). This is a highly distributed setting where each client has access to a small number of samples. We conducted our experiment using $\lambda = 0.08$. Our method recovered a support with 48 elements, with $39$ of them being in the ``centralized'' support. We report the following results:
	\begin{itemize}
		\item Recall: 97.5\%
		\item Precision: 81.2\%
		\item $F_1$-score: 88.7 \%
	\end{itemize}

	\paragraph{Case 2.} For the second experiment, we divided the dataset randomly into $52$ clients with each client containing $1000$ samples (except the last one which contains more to account for imbalance). This is a setting where we have only a few clients. We again conducted our experiment using $\lambda = 0.08$. Our method recovered a support with 49 elements, which included all of the elements in the ``centralized'' support. We report the following results:
	\begin{itemize}
		\item Recall: 100\%
		\item Precision: 81.6\%
		\item $F_1$-score: 90 \%
	\end{itemize}   
	
	We see that in both cases, we recovered almost all of the elements in the ``centralized'' support while making very few mistakes.


\begin{thebibliography}{10}
	
	\bibitem{bhagoji2019analyzing}
	Arjun~Nitin Bhagoji, Supriyo Chakraborty, Prateek Mittal, and Seraphin Calo.
	\newblock Analyzing federated learning through an adversarial lens.
	\newblock {\em International Conference on Machine Learning}, 2019.
	
	\bibitem{brendan2017aguera}
	McMahan~H Brendan, Moore Eider, Ramage Daniel, Hampson Seth, and Ag{\"u}era
	y~Arcas~Blaise.
	\newblock Communication-efficient learning of deep networks from decentralized
	data.
	\newblock In {\em Proceedings of the 20th International Conference on
		Artificial Intelligence and Statistics (AISTATS)}, 2017.
	
	\bibitem{he2018cola}
	Lie He, An~Bian, and Martin Jaggi.
	\newblock Cola: Decentralized linear learning.
	\newblock In {\em Advances in Neural Information Processing Systems}, pages
	4536--4546, 2018.
	
	\bibitem{konevcny2015federated}
	Jakub Kone{\v{c}}n{\`y}, Brendan McMahan, and Daniel Ramage.
	\newblock Federated optimization: Distributed optimization beyond the
	datacenter.
	\newblock {\em arXiv preprint arXiv:1511.03575}, 2015.
	
	\bibitem{konevcny2016federated}
	Jakub Kone{\v{c}}n{\`y}, H~Brendan McMahan, Felix~X Yu, Peter Richt{\'a}rik,
	Ananda~Theertha Suresh, and Dave Bacon.
	\newblock Federated learning: Strategies for improving communication
	efficiency.
	\newblock {\em arXiv preprint arXiv:1610.05492}, 2016.
	
	\bibitem{li2020federated}
	Tian Li, Anit~Kumar Sahu, Ameet Talwalkar, and Virginia Smith.
	\newblock Federated learning: Challenges, methods, and future directions.
	\newblock {\em IEEE Signal Processing Magazine}, 37(3):50--60, 2020.
	
	\bibitem{li2020fair}
	Tian Li, Maziar Sanjabi, and Virginia Smith.
	\newblock Fair resource allocation in federated learning.
	\newblock {\em International Conference on Learning Representations, ICLR},
	2020.
	
	\bibitem{mcdiarmid1989method}
	Colin McDiarmid.
	\newblock On the method of bounded differences.
	\newblock {\em Surveys in combinatorics}, 141(1):148--188, 1989.
	
	\bibitem{mohri2019agnostic}
	Mehryar Mohri, Gary Sivek, and Ananda~Theertha Suresh.
	\newblock Agnostic federated learning.
	\newblock {\em International Conference on Machine Learning}, 2019.
	
	\bibitem{ravikumar2011high}
	Pradeep Ravikumar, Martin~J Wainwright, Garvesh Raskutti, Bin Yu, et~al.
	\newblock High-dimensional covariance estimation by minimizing l1-penalized
	log-determinant divergence.
	\newblock {\em Electronic Journal of Statistics}, 5:935--980, 2011.
	
	\bibitem{smith2017federated}
	Virginia Smith, Chao-Kai Chiang, Maziar Sanjabi, and Ameet~S Talwalkar.
	\newblock Federated multi-task learning.
	\newblock In {\em Advances in Neural Information Processing Systems}, pages
	4424--4434, 2017.
	
	\bibitem{smith2017cocoa}
	Virginia Smith, Simone Forte, Chenxin Ma, Martin Tak{\'a}{\v{c}}, Michael~I
	Jordan, and Martin Jaggi.
	\newblock Cocoa: A general framework for communication-efficient distributed
	optimization.
	\newblock {\em The Journal of Machine Learning Research}, 18(1):8590--8638,
	2017.
	
	\bibitem{wainwright2015chapter}
	M~Wainwright.
	\newblock Chapter 2: Basic tail and concentration bounds, 2015.
	
	\bibitem{wainwright2009info}
	M.~J. Wainwright.
	\newblock Information-theoretic bounds on sparsity recovery in the
	high-dimensional and noisy setting.
	\newblock {\em IEEE Trans. Info. Theory}, 55:5728--5741, December 2009.
	
	\bibitem{wainwright2009sharp}
	Martin~J Wainwright.
	\newblock Sharp thresholds for high-dimensional and noisy sparsity recovery
	using l1-constrained quadratic programming (lasso).
	\newblock {\em IEEE transactions on information theory}, 55(5):2183--2202,
	2009.
	
	\bibitem{wang2020federated}
	Hongyi Wang, Mikhail Yurochkin, Yuekai Sun, Dimitris Papailiopoulos, and
	Yasaman Khazaeni.
	\newblock Federated learning with matched averaging.
	\newblock {\em International Conference on Learning Representations, ICLR},
	2020.
	
	\bibitem{yurochkin2019bayesian}
	Mikhail Yurochkin, Mayank Agarwal, Soumya Ghosh, Kristjan Greenewald,
	Trong~Nghia Hoang, and Yasaman Khazaeni.
	\newblock Bayesian nonparametric federated learning of neural networks.
	\newblock {\em International Conference on Machine Learning}, 2019.
	
	\end{thebibliography}
\end{document}